\documentclass[10pt,journal,cspaper,compsoc]{IEEEtran}
\usepackage{bm}
\usepackage{tabularx}

\usepackage{makecell}

\ifCLASSOPTIONcompsoc
  \usepackage[nocompress]{cite}
\else
\fi
\ifCLASSINFOpdf
    \usepackage[pdftex]{graphicx}
    \DeclareGraphicsExtensions{.pdf,.jpeg,.png}
\else
    \usepackage[dvips]{graphicx}
    \DeclareGraphicsExtensions{.eps}
\fi
\usepackage[cmex10]{amsmath}
\usepackage{algorithmic}
\usepackage{url}

\usepackage{multirow}
\usepackage{algorithm}

\usepackage{color}
\usepackage{amsmath}
\usepackage{amssymb}
\usepackage{mathrsfs}
\usepackage{booktabs}
\usepackage{comment}
\usepackage{multirow}
\usepackage{makecell,rotating}
\usepackage{array}
\usepackage{setspace}
\usepackage{framed}
\usepackage{bbm}

\newtheorem{theorem}{Theorem}
\newtheorem{corollary}{Corollary}
\usepackage[dvipsnames, svgnames, x11names]{xcolor}
\usepackage{extarrows}
\usepackage{soul}
\usepackage{hyperref}
\usepackage{cleveref}
\usepackage{svg}
\begin{document}
%
\title{The Interaction Bottleneck of Deep Neural Networks: Discovery, Proof, and Modulation
}

%
%

\author{Huiqi Deng\IEEEauthorrefmark{1}, 
Qihan Ren\IEEEauthorrefmark{1}, 
Zhuofan Chen,
Zhenyuan Cui,
Wen Shen,
Peng Zhang,
Hongbin Pei\IEEEauthorrefmark{2}, 
Quanshi Zhang\IEEEauthorrefmark{2}
\thanks{\IEEEauthorrefmark{1} Equal contribution.}
\thanks{\IEEEauthorrefmark{2} Corresponding author.}
\IEEEcompsocitemizethanks{\IEEEcompsocthanksitem 
Huiqi Deng, Zhuofan Chen,  Zhenyuan Cui, Peng Zhang, Hongbin Pei are with Xi'an Jiaotong University, Xi'an, China.
\IEEEcompsocthanksitem 
Qihan Ren and Quanshi Zhang are with Shanghai Jiao Tong University, Shanghai, China.
\IEEEcompsocthanksitem  Wen Shen is with Tongji University, Shanghai, China.}
}

%
%

\markboth{IEEE TRANSACTIONS ON PATTERN ANALYSIS AND MACHINE INTELLIGENCE}%
{Shell \MakeLowercase{\textit{et al.}}: Bare Demo of IEEEtran.cls for Computer Society Journals}
%


\IEEEcompsoctitleabstractindextext{%

\begin{abstract}
Understanding what kinds of cooperative structures deep neural networks (DNNs) can represent remains a fundamental yet insufficiently understood problem.
In this work, we treat interactions as the fundamental units of such structure and investigate a largely unexplored question:
how DNNs encode interactions under different levels of contextual complexity, and how these microscopic interaction patterns shape macroscopic representational capacity.
To quantify this complexity, we use multi-order interactions~\cite{zhang2020interpreting}, where each order reflects the amount of contextual information required to evaluate the joint interaction utility of a variable pair.
This formulation enables a stratified analysis of cooperative patterns learned by DNNs.
Building on this formulation, we develop a comprehensive study of interaction structure in DNNs. 
(i) We empirically \textit{discover} a universal \textit{interaction bottleneck}: 
across architectures and tasks, DNNs easily learn low-order and high-order interactions but consistently under-represent mid-order ones. 
(ii) We \textit{theoretically explain} this bottleneck by proving that mid-order interactions incur the highest contextual variability, yielding large gradient variance and making them intrinsically difficult to learn. 
(iii) We further \textit{modulate} the bottleneck by introducing losses that steer models toward emphasizing interactions of selected orders.
Finally, we connect microscopic interaction structures with macroscopic representational behavior: 
low-order-emphasized  models exhibit stronger generalization and robustness, whereas high-order-emphasized models demonstrate greater structural modeling and fitting capability.
Together, these results uncover an inherent representational bias in modern DNNs and establish interaction order as a powerful lens for interpreting and guiding deep representations.
\end{abstract}


}

\maketitle
\IEEEdisplaynotcompsoctitleabstractindextext
\IEEEpeerreviewmaketitle

\section{Introduction}
Deep neural networks (DNNs) have achieved remarkable empirical success, yet their inner mechanisms remain far from fully understood. 
A DNN’s prediction does not arise from isolated features; rather, it emerges from intricate cooperative structures—patterns describing how input variables jointly contribute to the outcome. 
Despite their central role, the nature of these cooperative structures and their influence on macroscopic behaviors remain unclear. 
This motivates two fundamental questions:
\begin{itemize}
    \item What kinds of cooperative structures are DNNs \textit{inherently capable or incapable} of representing?
    \item How do \textit{differences in these cooperative structures shape macroscopic behaviors} such as generalization, fitting ability, adversarial robustness, and structural modeling capacity?
\end{itemize}

\begin{figure*}[t]\centering
\includegraphics[width = 0.99 \textwidth]{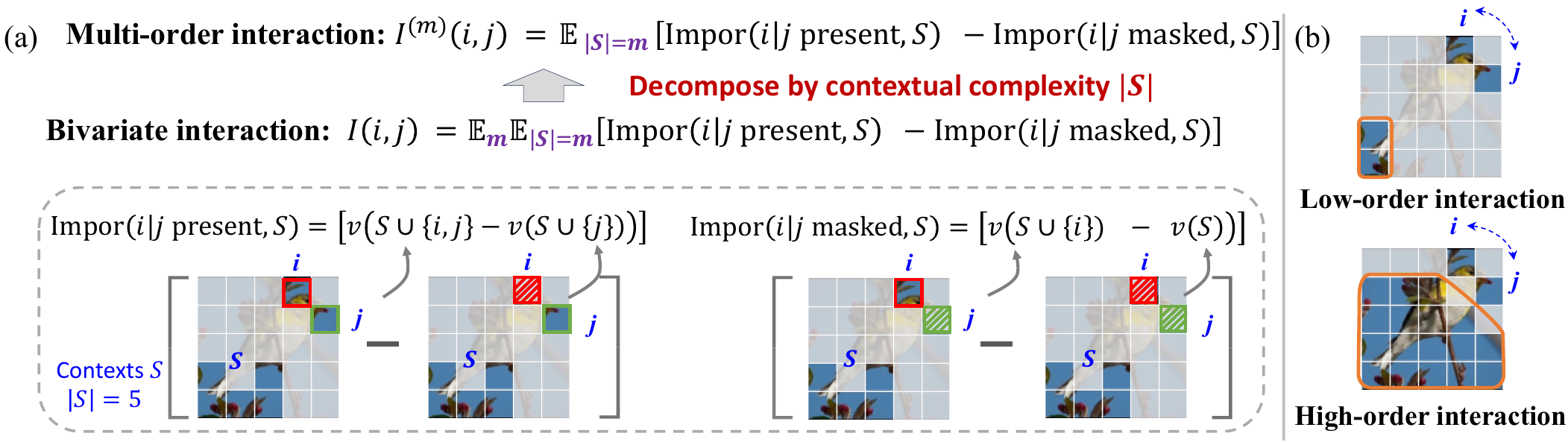}
\caption{(a) Shapley interaction index and multi-order interactions.
The bivariate interaction $I(i,j)$ quantifies the influence of the presence/absence of variable $j$ on the importance of variable $i$ across varying contexts.
By decomposing $I(i,j)$ by the contextual complexity, the $m$-th order interaction $I^{(m)}(i,j)$ represents the average interaction utility within contexts containing exactly $m$ variables ($|S| = m$). 
(b) Illustration of low-order and high-order interactions:
low-order interactions arise in simple, local contexts, 
whereas high-order interactions emerge when many contextual regions jointly influence the relationship between $i$ and $j$. 
}
\label{fig:figure1_added}
\vspace{-4pt}
\end{figure*}

\textbf{Multi-order interactions for characterizing cooperative structure.}
These questions call for a mechanistic investigation of interactions inside deep networks, as interactions constitute the most basic units of such cooperative structure.
To formally characterize them, we employ the Shapley Interaction Index (SII) $I(i,j)$~\cite{grabisch1999axiomatic}, which quantifies the nonlinear collaboration between two variables by measuring how the inclusion or exclusion of one variable changes the attributed importance of the other across different contexts (Fig.~\ref{fig:figure1_added}(a)).

While SII captures \textit{whether} two variables interact, it does not indicate \textit{how much contextual information} the interaction relies on—an essential dimension of representation complexity. 
To capture this dimension, we decompose each bivariate interaction $I(i,j)$ into multi-order components $I^{(m)}(i,j)$~\cite{zhang2020interpreting}, where the order $m$ denotes the size of contextual set accompanying $(i,j)$.
Each $I^{(m)}(i,j)$ isolates the interaction utility within contexts containing exactly $m$ variables, yielding a stratified view of cooperative structure: low-order interactions reflect simple, weakly contextual patterns; high-order interactions correspond to complex, strongly context-dependent relations; and mid-order interactions lie between them (Fig.~\ref{fig:figure1_added}(b)).

Prior work~\cite{zhang2020interpreting} shows that a DNN's prediction can be represented as a weighted sum over all multi-order interaction utilities across variable pairs.
Thus, multi-order interactions provide a natural bridge for connecting microscopic interaction patterns with macroscopic representational behavior such as generalization, adversarial robustness.

\begin{figure*}[t]\centering
\includegraphics[width = 1.0 \linewidth]{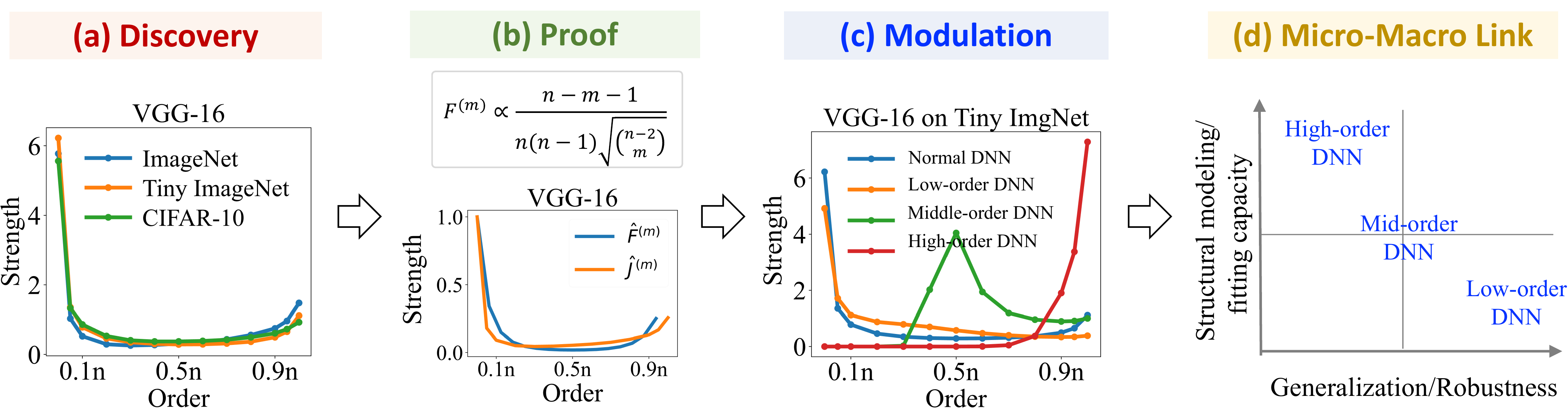}
\vspace{-8pt}
\caption{Overview. 
\textit{(a) Discovery:} Across datasets and architectures, DNNs consistently exhibit a universal interaction bottleneck—strong low-order and high-order interactions but weak mid-order interactions.
\textit{(b) Proof:} Our theoretical analysis characterizes how the learning strength $F^{(m)}$ of $m$-order interactions varies with contextual variability $\tbinom{n-2}{m}$, producing a curve that closely matches the empirical interaction distribution $J^{(m)}$.
\textit{(c) Modulation}: The proposed encouraging/suppressing losses offer explicit modulation of interaction orders, enabling DNNs to preferentially learn low-order, mid-order, or high-order interactions.
\textit{(d) Micro–Macro Link:} Micro-level interaction order shapes macro-level representation capability: high-order DNNs excel in structural modeling, low-order DNNs show superior generalization and robustness, and mid-order DNNs lie in between.
}
\label{fig:figure2_paper_overview}
\end{figure*}

\textbf{Discovering a universal interaction bottleneck.}
Using this multi-order decomposition, we observe a striking and previously overlooked regularity:
across CNNs, Vision Transformers, NLP Transformers, point-cloud networks, and MLPs, \textbf{\emph{DNNs strongly represent low-order and high-order interactions but exhibit markedly weaker representation of mid-order ones}}.
This \emph{interaction bottleneck} manifests as a characteristic U-shaped trend in the magnitudes of $I^{(m)}(i,j)$ across interaction orders (Fig.~\ref{fig:figure2_paper_overview}(a)).

This bottleneck reveals an inherent representation bias:
DNNs easily model simple, local low-order interactions and complex, global high-order interactions,
while mid-order interactions, neither clearly local nor fully global, receive much weaker representation.

\textbf{Proving the interaction bottleneck.}
To uncover the theoretical origin of this bottleneck, we analyze how the learnability of interactions varies across orders.  
We prove that the effective gradient signal for $I^{(m)}(i,j)$ is inversely related to its contextual variability $\tbinom{n-2}{m}$.

Consequently, low-order and high-order interactions arise under limited or nearly fixed contextual configurations, leading to low contextual variability and stable gradient directions, and are therefore readily learned.  
While mid-order interactions, correspond to maximal contextual variability, causing large gradient variance that hinder stable optimization. 
The resulting theoretical learning-strength curve closely  matches the empirical multi-order interaction distribution (Fig.~\ref{fig:figure2_paper_overview}(b)).

\textbf{Modulating the interaction bottleneck.}
Given the existence of this interaction bottleneck, a key question is whether DNNs’ interaction distributions can be explicitly modulated.
To this end, we introduce two complementary loss functions that respectively encourage or suppress interactions of designated orders.  
Both theoretical analysis and empirical evidence show that these losses effectively reshape the distribution of multi-order interactions, enabling DNNs to learn representations with targeted interaction distribution.

\textbf{From micro interactions to macro performance.}
Using this modulation mechanism, we construct models that predominantly learn low-order, mid-order, or high-order interactions and systematically evaluate their macroscopic representation capacities.

Low-order–biased models exhibit strong generalization and adversarial robustness but limited structural expressiveness and fitting capacity, whereas high-order–biased models offer richer structural modeling and fitting ability but show weaker generalization and robustness.  
Mid-order–biased models fall between these extremes.  
Together, these findings reveal a fundamental trade-off governed by interaction complexity:
(i) different interaction orders preferentially support different macroscopic capabilities, and
(ii) practical DNNs benefit from implicitly combining contributions across orders, thereby achieving a more balanced representational capacity.

\textbf{Contributions.} Our contributions are as follows:
\begin{itemize}
    \item We discover a universal interaction bottleneck phenomenon across architectures and modalities: DNNs strongly represent low-order and high-order interactions but consistently under-represent mid-order ones.
    \item We uncover the theoretical origin of this bottleneck by showing that mid-order interactions maximize contextual variability, resulting in maximal gradient variance and reduced learnability.
    \item We introduce two  complementary  loss functions and demonstrate that they provide explicit, order-level modulation of interaction distributions.
    \item We establish a micro–macro connection demonstrating how interaction order governs  macroscopic capabilities such as fitting, generalization, robustness, and structural expressiveness. 
\end{itemize}

\section{Related work}
Research related to our work spans three areas: the representational capacity of DNNs, the role of interactions in understanding model behavior, and the methodological extensions introduced in this journal version beyond our prior conference paper.

\subsection{Representation capacity of DNNs}
Evaluating and interpreting the representation capacity of DNNs has been a long-standing research topic and has attracted extensive attention since the emergence of deep learning.

Early studies primarily focused on expressivity, which refers to the DNN's ability to approximate complex function distributions.
This line of research explored the problem from multiple theoretical perspectives,
including analyzing the family of functions that DNNs can represent~\cite{arora2016understanding}, 
the maximum number of linear response regions in multilayer perceptrons (MLPs)~\cite{montufar2014number, pascanu2013number}, 
and the complexity of ReLU transformations~\cite{ren2022towards}. 
In addition, some studies examined how architectural parameters influence expressivity, revealing that the expressive power of DNNs grows exponentially with network depth~\cite{raghu2017expressive}, where in recurrent neural networks (RNNs), it depends on factors such as recurrent depth, feedforward depth, and skip coefficients~\cite{zhang2016architectural}.
These works advanced the understanding of DNNs’ theoretical capacity,
but they only characterized the optimal representation space under idealized assumptions, rather than the actual representation structures that emerge during training.

Beyond expressivity, subsequent studies have shifted toward analyzing generalization and adversarial robustness.
Researchers have sought to explain why deep models often achieve remarkably low generalization errors despite heavy overparameterization~\cite{xu2018understanding, chatterjee2022generalization}, and why they remain vulnerable to adversarial perturbations~\cite{ilyas2019adversarial}.
A growing body of evidence indicates that generalization is positively correlated with the flatness of the loss landscape~\cite{hochreiter1997flat, keskar2016large, li2018visualizing, neyshabur2017exploring}.
Furthermore, metrics such as stiffness~\cite{fort2019stiffness}, sensitivity~\cite{novak2018sensitivity}, and the CLEVER score~\cite{weng2018evaluating} have been proposed to quantify generalization and robustness.
However, these works still focus predominantly on macroscopic performance indicators, capturing the external behavior of DNNs rather than revealing their internal representation mechanisms.

In summary, most prior studies have characterized the representation capacity of DNNs from a macroscopic perspective, emphasizing theoretical capacity, generalization, and robustness.
In contrast, 
this work re-examines representation capacity from \textbf{\textit{a microscopic viewpoint}} by modeling how DNN encode interaction structures, investigating \textbf{\textit{how DNNs learn interactions of varying complexity, and establishing a micro–macro connection  between interaction complexity and macroscopic performance indicators.}}

\subsection{Interactions} 
Interactions among input variables encoded by DNNs have received growing attention in neural network interpretability.
A variety of metrics have been proposed to characterize how input variables collaborate or depend on each other, including the Shapley Interaction Index~\cite{grabisch1999axiomatic}, Shapley–Taylor Interaction Index~\cite{sundararajan2020shapley}, multi-order interaction~\cite{zhang2020interpreting}, and the Harsanyi dividend~\cite{ren2023defining}.
These studies use interactions to explain the internal decision-making behavior of DNNs.

Beyond explanation, interactions have also been used to study how DNNs’ representational capacity relates to interaction strength or interaction complexity.
For example, theoretical work interprets dropout as suppressing high-order interactions~\cite{lengerich2020dropout, zhang2020interpreting}, and interaction metrics have been applied to analyze model generalization~\cite{yao2023towards, zhou2024explaining}, adversarial transferability~\cite{wang2020unified, zhang2022proving}, adversarial robustness~\cite{wang2021interpreting, ren2021game}, and improving feature disentanglement~\cite{peebles2020hessian, tsang2018neural}.

Whereas prior works primarily use interactions to interpret model decisions, we \textbf{\textit{uncover an interaction bottleneck that identifies the interaction structures DNNs are intrinsically able or unable to represent.}}

\subsection{Significant extensions over conference paper}
A preliminary version of this work appeared at ICLR~\cite{deng2022discovering}.
While the conference version introduced the interaction bottleneck and provided initial theoretical analysis and empirical evidence, this journal version substantially expands and deepens the study in conceptual framing, theoretical grounding, empirical breadth, and overall presentation.
The major extensions are summarized below.

\textbf{(i) Extensive cross-domain validation of the interaction bottleneck (Section~\ref{sec:bottleneck}).}
We greatly broaden the empirical evidence for the interaction bottleneck by evaluating Vision Transformers on two datasets with five architectures (2×5), 3D point-cloud models on two datasets with five architectures (2×5), and NLP Transformers on three datasets with four architectures (3×4).
We further validate the bottleneck throughout training dynamics, showing that the pattern consistently emerges during learning.
This breadth of evidence goes far beyond the ICLR version, which examined only CNNs and MLPs.

\textbf{(ii) New analysis on fitting vs. generalization (Section~\ref{subsec:Comparative Study}).}
We extend the empirical study to fundamental macroscopic capacities, investigating how interaction complexity (order) shapes fitting and generalization, which were not examined in the ICLR version.
We show that low-order–biased models generalize well but fit poorly, whereas high-order–biased models fit strongly but generalize weakly, revealing an intrinsic trade-off governed by interaction order.

\textbf{(iii) Introduction of contextual variability as the underlying mechanism (Section~\ref{sec:proving and explaining}).}
We introduce the concept of \emph{contextual variability}, which quantifies how many distinct contextual configurations each interaction order accounts for.
This provides the missing physical and intuitive mechanism behind the interaction bottleneck: mid-order interactions have the highest contextual variability (Fig.~\ref{fig:prove_illustration}), yielding maximally unstable learning signals.
This offers a clear conceptual advance over the ICLR version, which presented algebraic derivations but did not identify the fundamental cause of the bottleneck.

\textbf{(iv) Conceptual reframing and semantic clarification (Section~\ref{sec:multi-order interactions} and \ref{subsec:Semantic Implications}).}
We reorganize and rewrite the introduction to multi-order interactions so that they arise naturally from the Shapley interaction framework, making the formulation more intuitive and theoretically grounded.
We further provide a semantic characterization of different interaction orders and clarify the physical meaning of the interaction bottleneck.
These conceptual enhancements were absent from the ICLR version and substantially improve clarity and interpretability.

\textbf{(v) Substantial improvements in exposition and organization.}
Beyond isolated section updates, we restructure the \emph{entire} paper—refining the motivation, clarifying the problem formulation, strengthening theoretical explanations, adding an overview and illustrative figures, and improving narrative flow.
These revisions make the journal version markedly more accessible and coherent than the conference version.

\begin{table}[t]
\renewcommand{\arraystretch}{1.1}
\setlength{\tabcolsep}{3.5pt}
\caption{Main notations used in this paper.}
\vspace{-4pt}
\centering
\begin{tabular}{cl}
\toprule
\textbf{Notation} & \textbf{Description} \\
\midrule
$v(\cdot)$       & Pre-trained DNN \\
$x$              & Input sample $[x_1, \cdots, x_n]^{\!T}$ \\
$N$              & Input variable index, $N = \{1, \cdots, n\}$ \\
$S$              & Subset of $N$, i.e., $S \subseteq N$ \\
$x_S$            & Sample with variables in $N \setminus S$ masked \\
$v(N)$           & Model output on the full input $x$ \\
$v(S)$           & Model output on the masked input $x_S$ \\
$I(i, j)$        & Bivariate interaction defined in Eq.~(\ref{eqn:bivariate_interaction}) \\
$I^{(m)}(i, j)$  & Multi-order interaction defined in Eq.~(\ref{eqn:definition of I^m(i,j)}) \\
$J^{(m)}$        & $m$-th order relative interaction strength \\
$\mu_i$          & Independent effect of variable $i$ \\
$L^+(r_1, r_2)$  & Promotion loss for $[r_1n, r_2n]$ orders \\
$L^-(r_1, r_2)$  & Suppression loss for $[r_1n, r_2n]$ orders  \\
\bottomrule
\end{tabular}
\label{tab:my_label}
\end{table}

\section{Multi-order Interactions: Definition and Semantic Roles} \label{section:bottleneck}
This section introduces the notion of multi-order interactions and explains how different interaction orders characterize distinct forms of variable collaboration.

\noindent (i) We formally define multi-order interactions, which measure the cooperative effect between two variables under a fixed contextual complexity (Sec.~\ref{sec:multi-order interactions});

\noindent (ii) We then analyze the semantic implications of different interaction orders, clarifying their meanings and functional roles in the decision process (Sec.~\ref{subsec:Semantic Implications}).

\subsection{Definition of Multi-order Interactions}\label{sec:multi-order interactions}
Understanding DNNs internally organize the relationships among input variables is essential for interpreting their decision processes.
This requires moving beyond independent attribution and examining how input variables \emph{interact} with one another.

Consider a pre-trained DNN $v(\cdot)$ and an input sample $x = [x_1, \dots, x_n]^T$ whose variables are indexed by $N = \{1,\dots,n\}$ (e.g., an image decomposed into $n$ patches). 
We denote by $v(N)$ the network output on the full input $x$\footnote{For example, in multi-category image classification, one may use $v(N) = \log \frac{P(\hat y = y^*|x)}{1-P(\hat y = y^{*}|x)}$ as a logit-based confidence score, where $P(\hat y = y^*|x)$ is the probability of predicting the correct label}.
To formally capture how a variable pair jointly contribute to this output, we begin by defining the notion of a bivariate interaction.\\

\noindent \textbf{Definition of Shapley bivariate interaction index.}
Originally introduced in cooperative game theory~\cite{grabisch1999axiomatic}, the Shapley bivariate interaction index has been adapted to deep learning as a way to quantify how two input variables collaborate.
Specifically, an interaction utility $I(i,j)$ between variables $i$ and $j$ measures how the presence of variable $j$ changes $i$'s contribution to the network output.
For instance, $I(i,j) = 0.1$ indicates that masking variable $j$ reduces the contribution of $i$ by $0.1$.

The main challenge lies in how to measure the contribution of $i$ under the presence or absence of $j$. 
A naive formulation compares the marginal effect $v(\{i,j\}) - v(\{j\})$ with $v(\{i\}) - v(\emptyset)$, where
$v(S)$ denotes the model output on a masked input $x_S$ with variables in $N \setminus S$ replaced by a baseline value\footnote{Typically chosen as the empirical mean of each variable.}. 
However, this approach ignores contextual dependencies: the interaction between $i$ and $j$ may vary substantially depending on other  involved contextual variables.

A more accurate measure of the interaction between $i$ and $j$ must account for the fact that their joint effect can vary across different contextual subsets.
Thus, for any contextual subset $S \subseteq N \setminus \{i,j\}$, we measure how the marginal contribution of $i$ changes when $j$ is added to the context $S$:
\begin{equation}
\begin{aligned}
\label{eqn:bivariate_interaction2}
     \Delta v(i,j,S) &= \underbrace{[v(S \cup \{i,j\}) - v(S \cup \{j\})]}_{i \text{'s contribution when $j$ is present given } S} \\ 
   &- \underbrace{[v(S \cup \{i\}) - v(S)]}_{i \text{'s  contribution when $j$ is absent given } S} 
    \end{aligned}
\end{equation}
The overall Shapley bivariate interaction $I(i,j)$ is  then obtained as the expectation
of $\Delta v(i,j,S)$ over all contextual subsets $S$, grouped by their subset size:
\begin{equation}
\begin{aligned}
\label{eqn:bivariate_interaction}
  I(i,j)   = \mathbb{E}_m \mathbb{E}_{S \subseteq N \setminus \{i,j\}, |S| = m} [\Delta v(i,j,S)] \\ 
    \end{aligned}
\end{equation}
Intuitively, $I(i,j)$ captures the average collaboration effect between $i$ and $j$ across different contextual conditions.
This pairwise formulation naturally serves as the basis for defining multi-order interactions, introduced next.\\

\begin{figure}[t]\centering
\includegraphics[width = 0.4999 \textwidth]{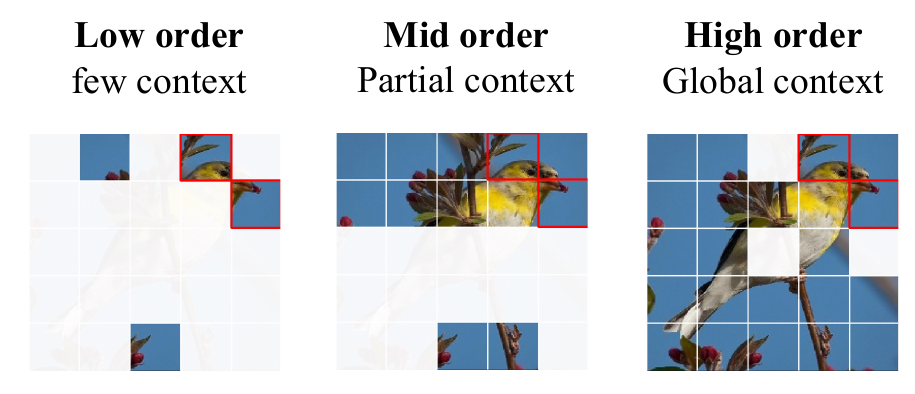}
\caption{
Illustration of low-order, mid-order, and high-order interactions.
The red bounding box marks the variable pair $(i,j)$, and the unmasked patches are treated as contextual variables.
Low-order interactions $I^{(m)}(i,j)$ use only a small context (e.g., $|S|=2$), mid-order interactions arise from intermediate context sizes (e.g., $|S|=10$), and high-order interactions rely on near-global context (e.g., $|S|=20$).
}
\label{fig:interaction_illustration}
\vspace{-4pt}
\end{figure}

\noindent \textbf{Definition of multi-order interaction.}
To characterize how the interaction utility between variables $i$ and $j$ changes with contextual complexity, Zhang et al.~\cite{zhang2020interpreting} introduced the multi-order interaction $I^{(m)}(i,j)$, which isolates their interaction utility  under contexts of fixed size $m$. Here, the order $m$ reflects the contextual complexity involved in the interaction.
Formally, the $m$-th order interaction is defined as:
\begin{equation}
\label{eqn:definition of I^m(i,j)}
\begin{aligned}
 I^{(m)}(i,j) & =  \mathbb{E}_{S \subseteq N \setminus \{i,j\}, |S| = m}[\Delta v(i,j,S)]  
\end{aligned}
\end{equation}
The Shapley bivariate interaction index $I(i,j)$ can then be written as the aggregation of multi-order interactions over orders:
\begin{equation}
    I(i,j) = \mathbb{E}_m \left[ I^{(m)}(i,j) \right]
\end{equation}
As Figure \ref{fig:interaction_illustration} shows, low-order interactions (small $m$) reflect how  variables $i, j$ collaborate under simple, limited contexts;
whereas high-order interactions (large $m$) reflect their joint effect under complex and massive contexts. \\

\noindent \textbf{Axiomatic foundations and links to existing interaction metrics.}
(i) \textit{Axiomatic foundations}: The multi-order interaction $I^{(m)}(i,j)$ satisfies five fundamental axioms, including linearity, nullity, commutativity, symmetry, and efficiency, establishing it as a mathematically well-founded measure~\cite{zhang2020game}.
(ii) \textit{Connections to established metrics}: Prior work~\cite{zhang2020interpreting, ren2021game} has shown  that multi-order interactions are theoretically linked to several widely used explanation metrics, including the Shapley value \cite{shapley1951notes, lundberg2017unified}, Shapley interaction index \cite{grabisch1999axiomatic}, Harsanyi interaction index \cite{harsanyi1958bargaining, ren2023defining}. 
These connections indicate that $I^{(m)}(i,j)$ is consistent with, rather than isolated from, established explanatory principles.
Details are provided in the Supplementary Materials.

These groundings give $I^{(m)}(i,j)$ a solid theoretical basis, lending credibility to the subsequent analyses.

\subsection{Semantic Roles of Low-order, Mid-order, and High-Order Interactions}
\label{subsec:Semantic Implications}
Building on the formal definition of multi-order interactions, we analyze their semantic roles.
For any variable pair $(i,j)$, multi-order interactions characterize how the contribution of $(i,j)$ cooperation evolves as the contextual complexity $m$ varies.
Thus, low-order, mid-order, and high-order interactions do not correspond to different variable pairs; rather, they reflect how the same pair $(i,j)$ behaves under minimal, partial, and near-global contextual conditions.
Figure~\ref{fig:interaction_illustration} illustrates a representative example.

\noindent \textbf{(i) Low-order interactions: local semantics under minimal context.}
Low-order interactions $I^{(m)}(i,j)$ are computed from extremely small contextual sets $S$ (very small $|S|$).
With almost no contextual cues available, the interaction is dominated by the intrinsic cooperative structure of $(i,j)$.
As shown in Figure~\ref{fig:interaction_illustration}, the $(i,j)$ region retains stable, easily recognizable semantics under weak contexts, revealing that low-order interactions encode local, fundamental, and context-independent semantic cues.

\noindent \textbf{(ii) High-order interactions: global semantics under near-complete context.}
High-order interactions $I^{(m)}(i,j)$ use a nearly complete contextual set ($|S|\approx n$).
As Figure~\ref{fig:interaction_illustration} shows, when global structure becomes available—such as body, pose, and background—it imposes strong semantic constraints that resolve the ambiguity of the local region $(i,j)$ (the bird head).
Accordingly, high-order interactions capture global, long-range semantic relations.

\noindent \textbf{(iii) Mid-order interactions: intermediate semantics under partial context.}
Mid-order interactions $I^{(m)}(i,j)$ rely on large yet incomplete  contextual sets (intermediate $|S|$).
Such partial contexts introduce more structural cues than low-order settings but still fail to reconstruct the global semantics, placing $(i,j)$ in an intermediate semantic regime. 
As shown in Figure~\ref{fig:interaction_illustration}, combinations such as “head + partial body” provide supplementary semantic hints, yet these cues vary markedly across different partial contexts, resulting in high semantic instability.
Therefore, mid-order interactions reflect  partially structured semantic relations whose cooperative effects fluctuate strongly with the specific contextual composition.

\textbf{Summary.} Low-order interactions yield stable local semantics, high-order interactions produce global semantics, and mid-order interactions lie between them and exhibit strong dependence on contextual changes.
These three semantic regimes, induced by differing contextual complexities, provide a principled lens for understanding how interaction orders shape the semantic organization of DNNs.

\begin{figure*}[t]\centering
\includegraphics[width = 0.999 \textwidth]{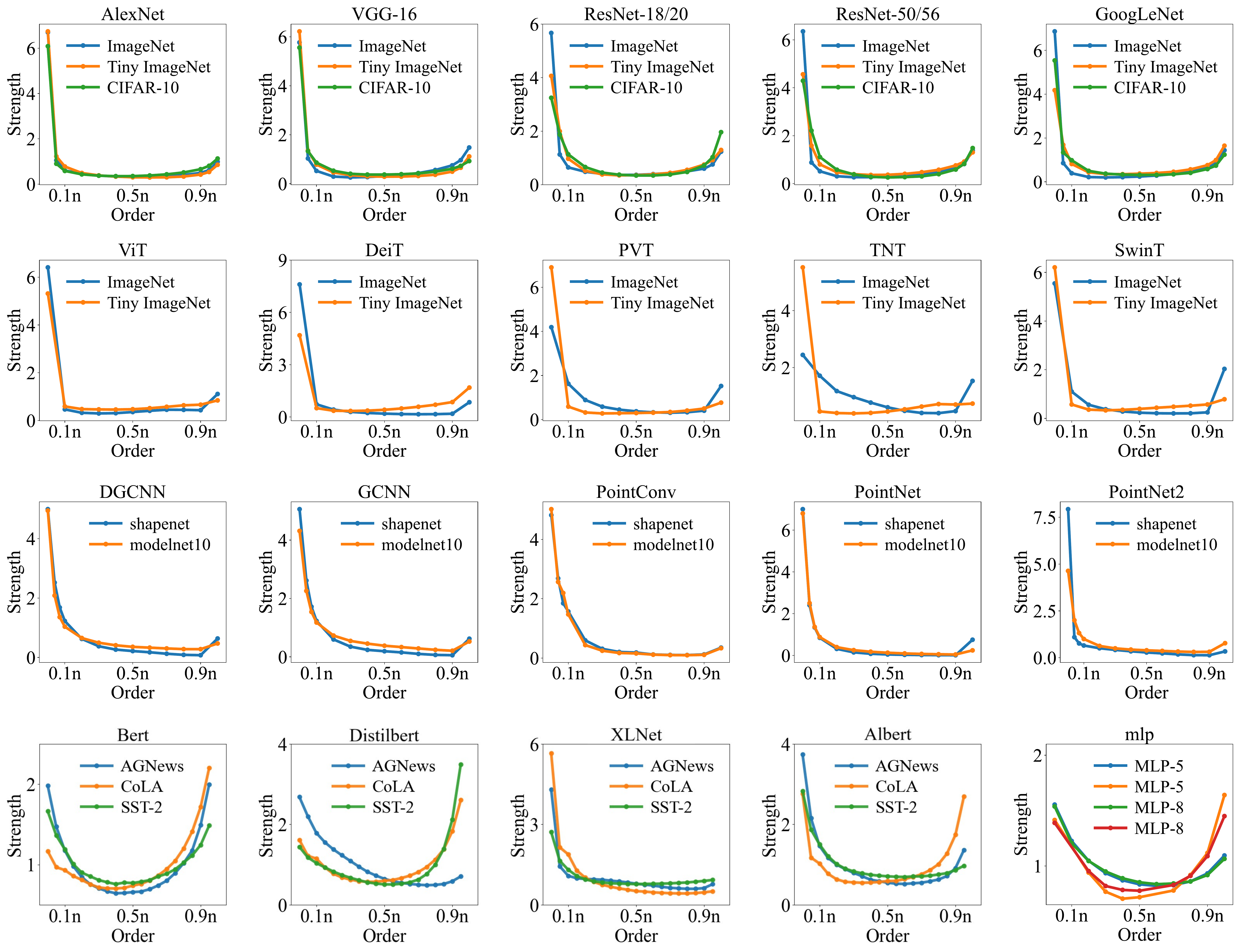}
\caption{Distributions of interaction strength $J^{(m)}$ across a wide range of DNNs, datasets, and architectures.
All curves consistently exhibit a characteristic \emph{interaction bottleneck}: DNNs emphasize low-order and high-order interactions while systematically downweighting mid-order ones.
}
\label{fig:interaction strength}
\end{figure*}

\section{Discovering the Interaction Bottleneck of DNNs}\label{sec:bottleneck}
Building on the above formulation, this section 

\noindent (i) establishes that a DNN’s prediction admits an exact decomposition over multi-order interactions, linking microscopic interaction patterns with macroscopic outputs (Sec.~\ref{subsec:Micro–Macro});

\noindent (ii) presents extensive cross-domain experiments discovering a universal \emph{interaction bottleneck}: DNNs model low-order and high-order interactions far more strongly than mid-order ones (Sec.~\ref{subsec:discover}); and

\noindent (iii) interprets this phenomenon through the intrinsic learning preferences of DNNs (Sec.~\ref{subsec:Semantic Implications of Interaction Bottleneck}).

\subsection{Micro–Macro Correspondence via Multi-order Interaction Decompositions}
\label{subsec:Micro–Macro}
Prior work~\cite{zhang2020interpreting} has shown that the prediction of a DNN can be decomposed exactly into the collective contributions of multi-order interaction patterns.
Formally, by the \textbf{\textit{efficiency property}} of multi-order interactions, the output on the full input $v(N)$ admits the following decomposition:
\begin{equation}\label{eqn:efficiency rewritten}
\begin{aligned}
v(N)  =  v(\emptyset) & + \sum_{i \in N} \mu_i +  \sum_{m = 0}^{n-2} w^{(m)} \cdot \bm{I}^{(m)} \\
\end{aligned}
\end{equation}
where $\mu_i = v({i}) - v(\emptyset)$ captures the individual contribution of variable $i$, and
$\bm{I}^{(m)} = \sum_{{i,j}\subseteq N} I^{(m)}(i,j)$ aggregates the interaction effects of all variable pairs under contextual complexity $m$, weighted by the coefficient $w^{(m)}$.

This decomposition establishes a clear micro–macro correspondence:
each term $\bm{I}^{(m)}$ serves as a microscopic reasoning component of order $m$, and the final prediction is obtained by their compositional integration across all orders.
Moreover, macroscopic representational behaviors—such as generalization, robustness—are influenced by how the model organizes interaction contributions across different orders.
\textit{In this sense, multi-order interactions provide a principled microscopic foundation for understanding the macroscopic representation capabilities of DNNs.}

\begin{figure*}[t]\centering
\includegraphics[width = 0.99\textwidth]{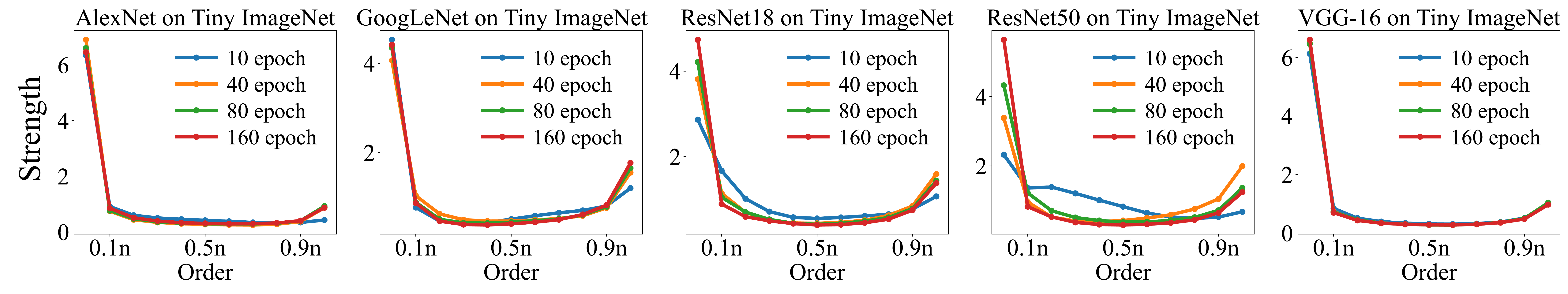}
\caption{
The distribution of interaction strength $J^{(m)}$ across different training epochs, showing that the interaction bottleneck persists throughout training.
}
\label{fig:dynamics}
\end{figure*}

\subsection{Discovering the Interaction Bottleneck}
\label{subsec:discover}
The efficiency property expresses the network output as a sum of interaction contributions across all orders.
Building on this decomposition, we examine how strongly a neural network represents interactions at each order and uncover a consistent bias in how these interactions are learned.\\

\noindent\textbf{Quantifying interaction strength across orders.}
To compare how strongly a network represents interactions of different complexities, we define the interaction strength at order $m$ for a given sample $x$ as
\begin{equation}
\begin{aligned}
     S^{(m)}(x) &= \mathbb{E}_{i,j} [\bm{|} I^{(m)}(i,j|x)\bm{|}]   
\end{aligned}
\end{equation}
which summarizes the magnitude of all pairwise interactions at that order.
A model-level measure $J^{(m)}$ is obtained by averaging over samples and normalizing across orders:
\begin{equation}
\begin{aligned}
      J^{(m)}  &=  \frac{\mathbb{E}_{x \in \Omega}[S^{(m)}(x)]}{\mathbb{E}_{m'}[\mathbb{E}_{x \in \Omega}[S^{(m')}(x)]]} 
\end{aligned}
\end{equation}
Here, $\Omega$ denotes the set of all input samples.
This measure allows us to directly compare the relative representation strength across interaction orders.\\

\noindent\textbf{Discovering the interaction bottleneck.}
By examining the distribution of interaction strength $J^{(m)}$, we uncover a striking and universal phenomenon across a wide range of architectures and tasks:
\begin{itemize}
\item \textbf{DNNs consistently encode stronger \emph{low-order} and \emph{high-order} interactions, while encode significantly weaker \emph{mid-order} interactions.}
\end{itemize}
As illustrated in Figure~\ref{fig:interaction strength}, the relative interaction strength $J^{(m)}$ typically shows a characteristic U-shape trend: it is high when the interaction order $m$ is either small (e.g., $m < 0.1n$) or large (e.g., $m > 0.9n$), but drops substantially around the mid-range (e.g., $m \approx 0.5n$).
We refer to this distinctive U-shaped distribution as the \textit{interaction bottleneck}.

We consistently observe this phenomenon across diverse architectures, modalities, and datasets:
\begin{itemize}
    \item \textbf{Vision CNNs}: five representative CNN models (AlexNet \cite{krizhevsky2012ImageNet}, VGG-16 \cite{simonyan2014very}, ResNet-18/20/50/56 \cite{he2016deep}, GoogLeNet \cite{szegedy2015going}) trained on CIFAR-10~\cite{krizhevsky2009learning}, Tiny-ImageNet~\cite{le2015tiny}, and ImageNet~\cite{ILSVRC15}.

    \item \textbf{Vision Transformers}: five representative Vision Transformer models, including ViT \cite{dosovitskiy2020image}, DeiT \cite{touvron2021training}, PVT \cite{wang2022pvt}, TNT \cite{han2021transformer}, and Swin Transformer \cite{liu2021swin}, trained on the Tiny-ImageNet, and ImageNet.
    
    \item  \textbf{3D Point Cloud Models}: five widely adopted architectures, i.e., DGCNN \cite{wang2019dynamic}, GCNN \cite{shen2021interpreting}, Pointconv \cite{wu2019pointconv}, PointNet \cite{qi2017pointnet}, and PointNet++ \cite{qi2017pointnet++}, trained on the ModelNet10 \cite{wu20153d} and ShapeNet \cite{yi2016scalable} benchmarks for 3D shape classification.
    
    \item \textbf{NLP Transformers}: four popular Transformer-based NLP models, including BERT \cite{devlin2019bert}, DistilBERT \cite{sanh2019distilbert}, XLnet~\cite{yang2019xlnet}, and ALBERT \cite{lan2019albert}, on SST-2 \cite{socher2013recursive}, AG News \cite{zhang2015character}, and CoLA \cite{warstadt2019neural}. 
    
    \item \textbf{Tabular MLPs}: two multilayer perceptrons (MLPs) with 5 and 8 layers, trained on the UCI census and UCI commercial datasets. 
\end{itemize}
Across all these settings, the interaction bottleneck appears consistently, demonstrating its remarkable generality and robustness. 

Moreover, the interaction bottleneck is not exclusive to fully trained models, but rather \textbf{persists consistently throughout the entire training process}.
As shown in Figure~\ref{fig:dynamics}, the characteristic U-shaped distribution of $J^{(m)}$ emerges early (e.g., by epoch 10) and remains highly stable over time, up to final epochs. 
This suggests that the interaction bottleneck is an intrinsic property of DNN learning dynamics,  rather than a byproduct of convergence.
Implementation details are provided in the supplementary materials.
\\



\subsection{Semantic Implications}
\label{subsec:Semantic Implications of Interaction Bottleneck}
The interaction bottleneck reveals an inherent inductive bias in how DNNs utilize interactions of different orders. Specifically, 

\noindent (i) DNNs preferentially capture low-order interactions, which reflect simple and context-light dependencies that generalize reliably across samples.
    
\noindent (ii) DNNs also reliably learn high-order interactions, which arise under rich contexts and encode coherent global relationships.
    
\noindent (iii) In contrast, DNNs downweight mid-order interactions, whose semantic roles fluctuate across partially informative contexts.

\begin{figure*}[t]\centering
\includegraphics[width = 0.99 \textwidth]{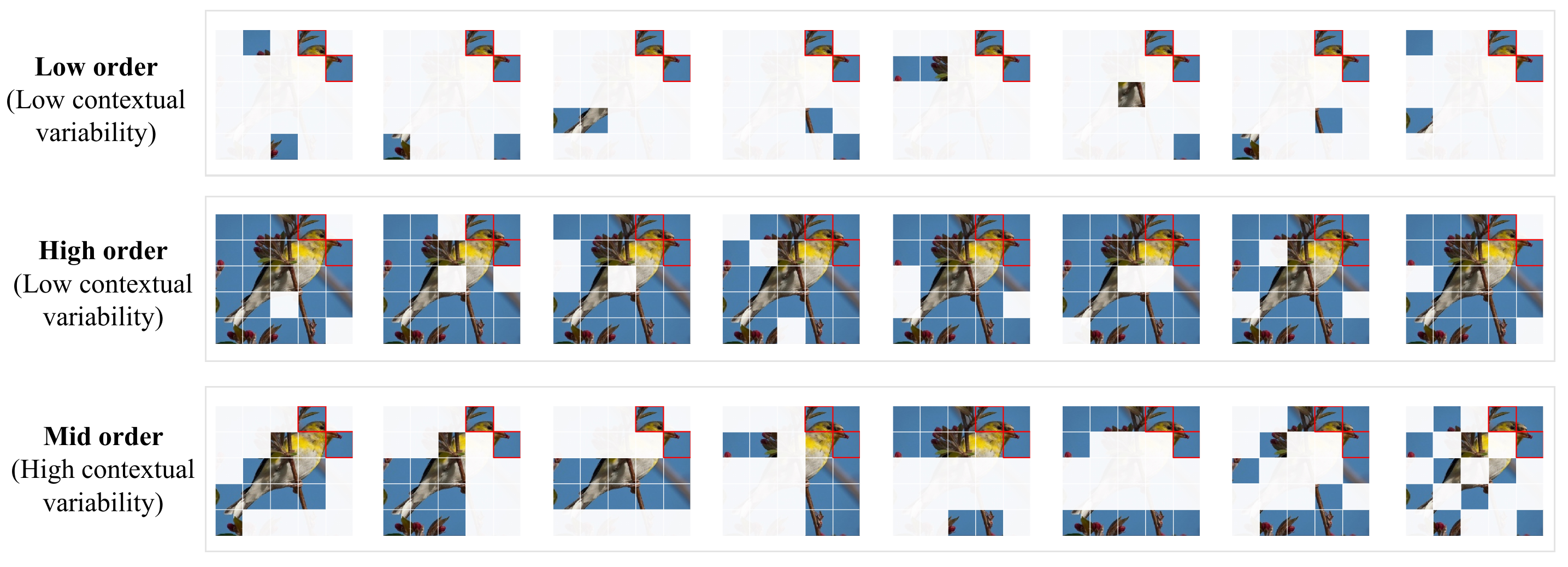}
\vspace{-4pt}
\caption{
Contextual variability across interaction orders for a fixed variable pair $(i,j)$ (highlighted in red box).
Low-order and high-order interactions occur under contexts with limited variation,
whereas mid-order interactions arise under diverse and heterogeneous partial contexts (e.g., tail-only regions, partial-body fragments, or background patches), resulting in much higher variability.
This disparity in contextual variability explains why DNNs struggle to learn mid-order interactions.
}
\label{fig:prove_illustration}
\vspace{-4pt}
\end{figure*}

\section{Proving and Explaining the Interaction Bottleneck}\label{sec:proof}
In this section, we uncover the theoretical origin of the universally observed interaction bottleneck from a learning perspective:
\textit{Why do DNNs reliably learn low-order and high-order interactions, yet systematically underrepresent mid-order ones?}

We identify that this bottleneck originates from the systematic differences in \textbf{contextual variability} across interaction orders, with the mid-order interactions exhibiting the highest variability and consequently the greatest learning difficulty.

\subsection{Quantifying Learning Strength across Orders}    
To theoretically explain the interaction bottleneck,
we quantify how strongly a DNN learns interactions of different orders during training.

To this end, we decompose the parameter update in each training step into components associated with interactions of different orders.
Specifically, let $W \in \mathbb{R}^K$ denote the parameters of the network.  
The parameter update $\Delta W$ is given by:
\begin{equation}\label{eqn:DeltaW}
\Delta W = - \eta \frac{\partial L}{\partial W} = - \eta \frac{\partial L}{\partial v(N)} \cdot \frac{\partial v(N)}{\partial W}
\end{equation}
where $L$ is the loss and $\eta$ is the learning rate.

Using the efficiency property in Eq.~(\ref{eqn:efficiency rewritten}), the network output $v(N)$ can be decomposed into a sum of multi-order interaction terms $I^{(m)}(i,j)$.
Consequently, the parameter update $\Delta W$ can be decomposed into gradient components associated with these multi-order interactions:
\begin{equation}\label{eqn:efficiency-gradient}
\begin{aligned}
 \Delta W &= \Delta W_U + \sum\limits_{m = 0}^{n-2} \sum\limits_{i,j \in N, i \neq j} \Delta W^{(m)}(i,j) 
\end{aligned}
\end{equation}
where 
\begin{equation}\label{eqn:efficiency-gradient-subeqn1}
\begin{aligned}
\Delta W_U & \triangleq - \eta \frac{\partial L}{\partial v(N)} \frac{\partial v(N)}{\partial U} \frac{\partial U}{\partial W} \\
\Delta W^{(m)}(i,j) & \triangleq - \eta \frac{\partial L}{\partial v(N)} \frac{\partial v(N)}{\partial I^{(m)}(i,j)} \frac{\partial I^{(m)}(i,j)}{\partial W}
\end{aligned}
\end{equation}
with $U = v(\emptyset) + \sum_{i \in N} \mu_i$ denoting the aggregated independent effect of all input variables.

According to the above decomposition, the parameter update $\Delta W$ separates into two distinct learning contributions:
$\Delta W_U$ captures the update associated with learning the independent effect $U$,
whereas $\Delta W^{(m)}(i,j)$ captures the update associated with learning each multi-order interaction $I^{(m)}(i,j)$ between variable pairs $(i,j)$.
Then, we define the \textbf{\textit{learning strength}} of $m$-order interactions as the expected $L_2$ norm of the corresponding gradient component:
\begin{equation}\label{eqn:learning-strength}
F^{(m)} \triangleq E_{i,j}\|\Delta W^{(m)}(i,j)\|_2
\end{equation}
A larger $F^{(m)}$ indicates that $m$-order interactions exert a stronger influence on the parameter update.

\subsection{Proving and Explaining Interaction Bottleneck as a Consequence of Contextual Variability}
\label{sec:proving and explaining}
In this subsection, we derive a closed-form expression for the learning strength of multi-order interactions and show that it closely follows a characteristic U-shaped curve, thereby theoretically confirming the interaction bottleneck in DNNs.
We further demonstrate that this U-shaped behavior arises fundamentally from the uneven distribution of \textbf{contextual variability} across interaction orders.\\

\noindent\textbf{Definition of contextual variability.}
For an $m$-order interaction, the \textit{contextual variability} is defined as the number of possible contextual sets at that order:
\begin{equation}
u(m) = \tbinom{n - 2}{m}
\end{equation}
This quantity measures how many distinct subsets of the remaining $(n-2)$ variables can serve as context.
Because $u(m)$ peaks at $m \approx (n-2)/2$, mid-order interactions exhibit the highest contextual variability, whereas low-order and high-order interactions correspond to much lower variability.
\vspace{3pt}


\begin{figure*}[t]\centering
\includegraphics[width = 1.0 \textwidth]{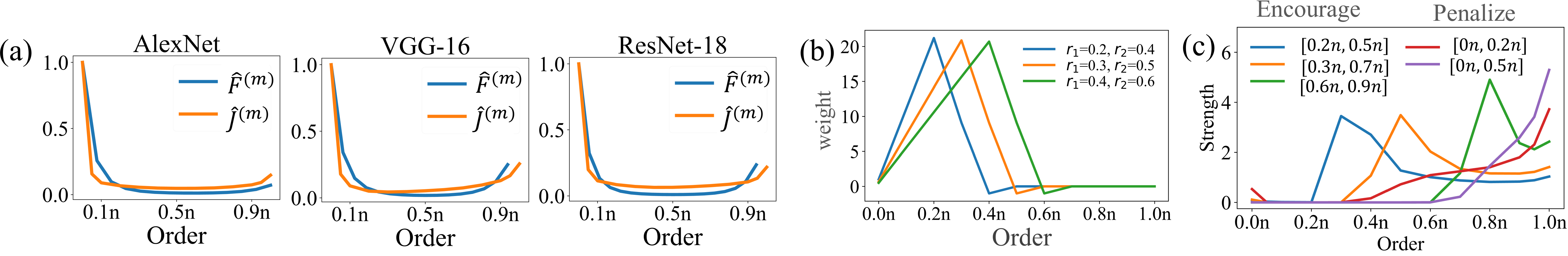}
\caption{
(a) Theoretical learning-strength curves $\hat F^{(m)}$ derived from Theorem~\ref{thm: interaction bottleneck} closely match the empirical interaction-strength distributions $\hat J^{(m)}$ across different networks, confirming the predictive power of the theory.
(b) Order-wise decomposition weights of $\Delta u(r_1, r_2)$.
(c) Interaction-strength distributions of DNNs trained with the proposed encouraging  loss $L^{+}$ (specific orders) and penalizing loss $L^{-}$, demonstrating controllable modulation of interaction complexity.
}
\label{fig:simulation and weight}
\end{figure*}

\noindent\textbf{Proof of the interaction bottleneck.} 
We now derive a closed-form expression for the learning strength of multi-order interactions, from which the interaction bottleneck follows immediately.
\vspace{5pt}

\begin{theorem}
\label{thm: interaction bottleneck}
\textit{(Proof in supplementary materials) 
Assume that for any order $m$,
$\mathbb{E}_{i,j} \mathbb{E}_{|S|=m}[\frac{\partial \Delta v(i,j,S)}{\partial W}]$ $= \bm{0}$\footnote{The zero-mean assumption is discussed and empirically validated in the supplementary materials.}.
Let $\sigma^2$ denote the variance of each coordinate of $\frac{\partial \Delta v(i,j,S)}{\partial W}$.
Under this assumption, the learning strength of order-$m$ interactions satisfies
\begin{equation}
\begin{aligned}\label{eqn:learning strength}
F^{(m)}  =   \frac{C \sigma  \cdot (n-m-1)}{n(n-1)} \frac{1}{\sqrt{u(m)}}
\end{aligned}
\end{equation}
where $C = \sqrt{K \eta \frac{\partial L}{\partial v(N)}} $ is a term independent of the order $m$, and $K$ denotes the dimensionality of $W$.} 
\end{theorem}
\vspace{3pt}

\begin{corollary}
\label{corollary1}
\textit{The learning strength of multi-order interactions decreases as contextual variability increases.}
\end{corollary}
\vspace{3pt}

Theorem~\ref{thm: interaction bottleneck} and Corollary~\ref{corollary1} show that the learning strength of $m$-order interactions, $F^{(m)}$, is inversely proportional to their contextual variability $u(m)$.
Consequently, low-order and high-order interactions, where contextual variability is small, receive stronger learning signals, while mid-order interactions—where contextual variability attains its maximum—receive substantially weaker learning signals.
This inverse dependence yields a characteristic U-shaped learning-strength curve, formally establishing the interaction bottleneck phenomenon.
\\

\noindent\textbf{Intuitive explanation.}
Fig.~\ref{fig:prove_illustration} provides intuition for why different interaction orders exhibit different levels of learnability 
by illustrating how contextual variability systematically changes with $m$, and how this variability affects the stability of the gradients associated with each order.
\begin{itemize}
    \item \textbf{Low-order and high-order interactions} are evaluated under either very small contexts or nearly complete ones, where the set of admissible conditioning configurations is inherently limited. As shown in Fig.~\ref{fig:prove_illustration}, these contexts vary only minimally across conditioning sets, and such stability translates into consistent gradient directions that the model can aggregate reliably.

    \item \textbf{Mid-order interactions}, in contrast, arise under an extremely large number of possible partial contexts, reaching up to $\tbinom{n-2}{n/2}$. As illustrated in Fig.~\ref{fig:prove_illustration}, these partial contexts differ widely—highlighting background-only regions, partial body segments, or mixtures of unrelated patches. 
    This high contextual variability leads to highly inconsistent gradient directions across conditioning sets, substantially weakening the learnability of mid-order interactions.
\end{itemize}

Thus, the interaction bottleneck arises naturally from the uneven distribution of contextual variability across interaction orders:
DNNs learn interactions most effectively when their contextual conditions are stable (low orders and high orders) and struggle particularly when the contextual space becomes highly variable (mid orders).

\subsection{Agreement between Theoretical Prediction and Empirical Observations}
To validate our theory, we use the analytically derived learning strength $F^{(m)}$ to simulate the empirical interaction strengths $J^{(m)}$ observed in real-world DNNs, showing strong agreement between the two.

To enable a fair comparison, we normalize both quantities as $\hat{F}^{(m)} = F^{(m)} / F^{(0)}$ and $\hat{J}^{(m)} = J^{(m)} / J^{(0)}$, providing a common reference point with $\hat{F}^{(0)} = \hat{J}^{(0)} = 1$.
Moreover, as DNNs usually learn highly redundant feature representations, their effective latent dimensionality $n'$ is typically much smaller than the number of input variables $n$.
We thus compute $\hat{F}^{(m)}$ using $n'$ (with $n' << n$), which yields a theory curve that better aligns with the behavior of practical DNNs.

As shown in Figure~\ref{fig:simulation and weight}(a), the theoretical curve $\hat{F}^{(m)}$ closely matches the empirical distribution $\hat{J}^{(m)}$, confirming that the derived theory faithfully captures the interaction bottleneck observed in practice.

\begin{figure*}[t]\centering
\includegraphics[ width=0.95\textwidth]{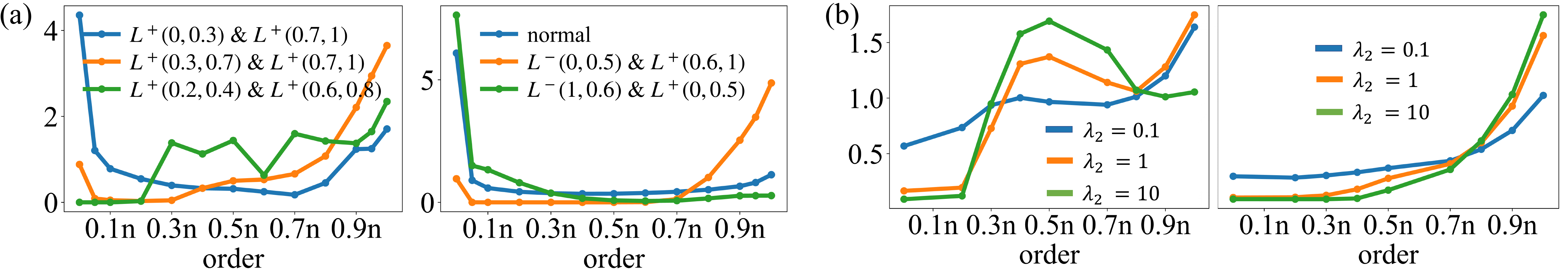}
\vspace{-0.2cm}
\caption{
(a) Interaction-strength distributions $J^{(m)}$ for models trained on CIFAR-10.
Left: applying multiple encouraging losses $L^{+}$.
Right: combining encouraging $L^{+}$ and penalizing $L^{-}$ losses.
(b) Effect of hyperparameters $\lambda_1$ and $\lambda_2$.
Left: models trained with $\lambda_{2}=0$ to encourage orders in $[0.3n,0.7n]$, varying $\lambda_{1}\in \{0.1,1,10\}$.
Right: models trained with $\lambda_{1}=0$ to penalize orders in $[0,0.5n]$, varying $\lambda_{2}\in \{0.1,1,10\}$.
}
\label{fig:control_interactions}
\end{figure*}

\section{Modulating Interaction Orders}
\label{subsection:proposedloss}
Having uncovered and theoretically explained the ubiquitous interaction bottleneck, we now ask whether a model’s learning emphasis across interaction orders can be explicitly controlled.
To this end, we introduce two modulation losses (Sec.~\ref{subsec:losses}) and demonstrate their effectiveness in steering a model toward desired interaction structures (Sec.~\ref{subsec:losseffective}).
Using these losses, we instantiate three representative DNN variants—favoring low-order, mid-order, and high-order interactions, and examine how these distinct interaction regimes influence macroscopic properties such as generalization, robustness, and structural modeling capability (Sec.~\ref{subsec:Comparative Study}).

\subsection{Losses for Interaction Order Modulation}
\label{subsec:losses}
We introduce two simple yet effective losses that explicitly modulate the interaction structure—one that encourages interactions within a targeted order range and another that suppresses them.

The key idea is to construct a controllable signal, $\Delta u(r_1,r_2)$, which aggregates interactions within the order range $[r_1 n, r_2 n]$. 
The encouraging loss forces the model to classify using this signal, whereas the suppressing loss discourages reliance on it.\\

\noindent \textbf{Core component $\Delta u(r_1,r_2)$.}
To isolate the contribution of specific interaction orders, we define the output-difference term:
\begin{equation}\label{eqn:component Delta u}
\begin{aligned}
    \!\!\! \Delta u(r_1,r_2) \triangleq \mathop{\mathbb{E}}\nolimits_{\emptyset \subseteq S_1 \subsetneq S_2 \subseteq N \atop 
     |S_1| = r_1n, |S_2| = r_2n} [v(S_2) -  \frac{r_2}{r_1} \cdot v(S_1)],
\end{aligned}
\end{equation}
where the expectation is taken over subset pairs $(S_1, S_2)$ satisfying $\emptyset \subseteq S_1 \subsetneq S_2 \subseteq N$, and $0 \le r_1 < r_2 \le 1$. 
Intuitively, the difference between $v(S_2)$ and  the rescaled  $v(S_1)$ cancels out all interactions shared by both subsets, leaving only the interactions that are activated when context grows from the size $r_1n$ to $r_2n$. This intuition is formalized in the following theorem.

\begin{theorem}
\label{thm:Delta}
\textit{(Proof in supplementary materials) The output change $\Delta u(r_1,r_2)$ can be expressed as follows:
\begin{equation}
\begin{aligned}\nonumber
     \Delta u(r_1, r_2) = (1 - \frac{r_2}{r_1}) v(\emptyset) +  \sum\limits_{m=0}^{n-2} \sum\limits_{i,j\in N
     \atop i \neq j}  \tilde w^{(m)} \cdot I^{(m)}(i,j) 
\end{aligned}
\end{equation}
where
\begin{equation}\nonumber
         \tilde w^{(m)} =
\left\{
            \begin{array}{ll}
             \frac{(\frac{r_2}{r_1}-1)(m+1)}{n(n-1)}, & m \leq r_1n - 2  \\
              \frac{(r_2n - m - 1)}{n(n-1)},   & r_1n - 2  < m \leq r_2n - 2 \\
             0,    &   r_2n - 2 < m  \leq n-2
             \end{array}
\right.
\end{equation}
}
\end{theorem}

Theorem~\ref{thm:Delta} shows that $\tilde w^{(m)}$ concentrates around $m=r_1n$ and vanishes beyond $m=r_2n$, meaning that $\Delta u(r_1,r_2)$ primarily captures interactions up to order $r_2n$, with emphasis around $r_1n$ (See Figure~\ref{fig:simulation and weight}(b)).
This provides the theoretical basis for selectively promoting or penalizing specific interaction orders.\\

\noindent\textbf{Losses for encouraging and suppressing specific interaction orders.} 
To regulate how a DNN utilizes interactions in a chosen order range, we design two complementary losses:
an encouraging loss $L^+(r_1,r_2)$ and a suppressing loss $L^-(r_1,r_2)$.

\textit{Encouraging loss.}
The encouraging loss $L^+(r_1,r_2)$ trains the model to perform classification using only the interaction signal contained in the order range $[r_1 n, r_2 n]$, thereby amplifying the learning of interactions within this interval:
\begin{equation}\nonumber
\begin{aligned}
& L^+(r_1,r_2) = \mathbb{H}(P(y), \hat P(y|\Delta u(r_1,r_2))) \\ 
\end{aligned}
\end{equation}
For each sample $x$ and class $c$, we define an \emph{interaction-based logit} $\Delta u_c(r_1,r_2 \mid x)$ that aggregates only those interaction contributions whose contextual sizes fall within $[r_1 n, r_2 n]$.
Following Eq.~(\ref{eqn:component Delta u}), it is computed by averaging differences of masked-subset logits $v_c(S|x)$:
\begin{equation}\nonumber
    \begin{aligned}
\!\!\! \Delta u_c(r_1,r_2|x)  \triangleq \mathop{\mathbb{E}}\nolimits_{\emptyset \subseteq S_1 \subsetneq S_2 \subseteq N \atop 
     |S_1| = r_1n, |S_2| = r_2n}   [v_c(S_2|x) -  \frac{r_2}{r_1} v_c(S_1|x)] 
     \end{aligned}
\end{equation}
This $\Delta u_c(r_1,r_2\mid x)$ serves as the effective logit when the model is restricted to the specified interaction interval.
The predicted probability $\hat P(y \mid \Delta u(r_1,r_2\mid x))$ is then obtained by applying a softmax over the class-wise values $\{\Delta u_c(r_1,r_2\mid x)\}_{c=1}^C$.

\textit{Suppressing loss.} 
The suppressing loss $L^{-}(r_1,r_2)$ reduces the model’s reliance on interactions within the range $[r_1 n, r_2 n]$ by maximizing the entropy of the prediction based solely on the interaction signal:
\begin{equation}\nonumber
\begin{aligned}
    & L^{-}(r_1,r_2) = - \mathbb{H}(\hat P(y|\Delta u(r_1,r_2|x))) 
\end{aligned}
\end{equation}
The negative entropy term penalizes confident predictions derived from $\Delta u(r_1,r_2)$thereby reducing their influence in the learned representation.

\textit{Overall loss.}
By combining the two modulation losses, we can obtain a model biased toward any desired interaction-order distribution.
\begin{equation}\label{eqn:new_loss}
 L  = L_{\textrm{cls}} + \lambda_1 L^{+}(r_1,r_2) +\lambda_2 L^{-}(r_1',r_2')
\end{equation}
where $L_{\textrm{cls}}$ is the standard classification cross-entropy, and $\lambda_1, \lambda_2 \ge 0$
control the modulation strength.

\begin{figure*}[t]\centering
\includegraphics[width= 1.0 \textwidth]{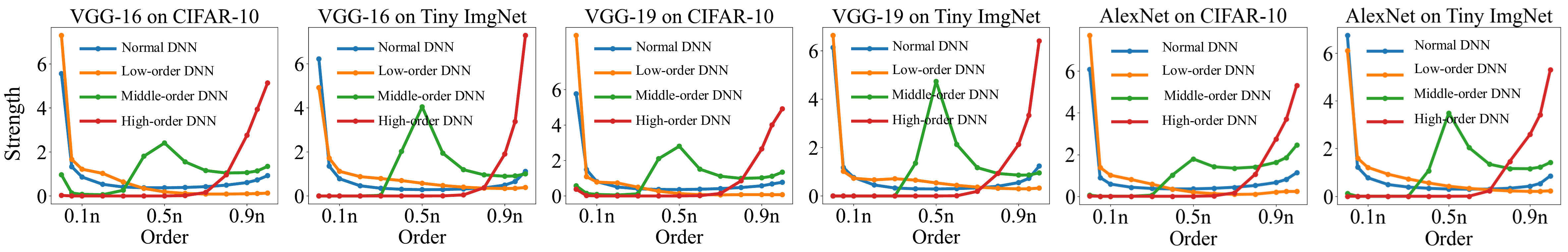}
\caption{
Distributions of the interaction strength $J^{(m)}$ for four representative types of DNNs.
Low-order, mid-order, and high-order DNNs each concentrate their interaction strength on the corresponding interaction orders,
while normal DNNs show the characteristic U-shaped interaction pattern.
}
\label{fig:investigate_appendix}
\end{figure*}

\begin{table*}[t]
\renewcommand{\arraystretch}{1.1}
\centering
\caption{
Comparison of models biased toward low-order, mid-order, and high-order interactions across four key dimensions.
}
\begin{tabular}{lcccc}
\toprule
\textbf{Model} & \textbf{Structural Modeling} & \textbf{Fitting Ability}  & \textbf{Generalization} & \textbf{Adversarial Robustness} \\ 
\midrule
Low-order DNNs      & Low   & Low  & High   & High    \\ 
Mid-order DNNs   & Medium & Medium  & Medium & Medium  \\ 
High-order DNNs     & High  & High   & Low    & Low     \\ 
\bottomrule
\end{tabular}
\label{tab:comprehensive_comparison_multi_order}
\vspace{-4pt}
\end{table*}

\subsection{Effectiveness of Proposed Losses}
\label{subsec:losseffective}
We now empirically validate  the effectiveness of the proposed loss, i.e., they indeed provide fine-grained modulation over interaction orders. \\

\noindent \textbf{Single loss effect.}
We first examine the individual effects of the encouraging and suppressing losses.
For $L^+(r_1, r_2)$, we enable only the encouraging term ($\lambda_1 = 1, \lambda_2 = 0$), and train three AlexNet models on Tiny-ImageNet using $(r_1, r_2)$ $\in $ $\{(0.2, 0.5), (0.3, 0.7), (0.6, 0.9)\}$.
For $L^-(r_1, r_2)$, we activate only the suppressing term ($\lambda_1 = 0$, $\lambda_2 = 1$), and train two additional AlexNet models with $(r_1, r_2) = (0, 0.2)$ and $(0, 0.5)$.

As shown in Fig.~\ref{fig:control_interactions}(b), $L^+(r_1, r_2)$ markedly boosts the interaction strength $J^{(m)}$ within the target order interval $[r_1n, r_2n]$, whereas $L^-(r_1, r_2)$ effectively suppresses interactions in the corresponding range. 
These results demonstrate that each loss term modulates interaction orders exactly as intended.\\

\noindent \textbf{Effect of combining multiple losses.}
We next test whether combining multiple terms $L^+$ and $L^-$ yields more flexible interaction control.
Five configurations are evaluated by training AlexNet models on CIFAR-10,
using either pairs of encouraging losses or mixed encouraging–suppressing combinations:
(i) $L^{+}(0,0.3)$ and $L^{+}(0.7,1.0)$;
(ii) $L^{+}(0.3,0.7)$ and $L^{+}(0.7,1.0)$;
(iii) $L^{+}(0.2, 0.4)$ and $L^{+}(0.6,0.8)$;
(iv) $L^{+}(0.6,1.0)$ and $L^{-}(0,0.5)$;
(v) $L^{+}(0,0.5)$ and $L^{-}(0.6,1.0)$.

The resulting distributions of $J^{(m)}$ are shown in Fig.~\ref{fig:control_interactions}(a).
These results demonstrate that combining multiple losses enables substantially more flexible and fine-grained control of interaction orders.
For instance, $L^+(0,0.3)$ with $L^+(0.7,1.0)$ boosts both low and high orders, while $L^+(0.6,1.0)$ with $L^-(0,0.5)$ enhances higher orders and suppresses lower ones.
Notably, multi-loss configurations can generate interaction distributions unattainable via single-loss setups—for example, combining $L^+(0.2,0.4)$ and $L^+(0.6,0.8)$ produces a clear bimodal pattern, whereas single losses typically yield unimodal curves.
\\

\noindent \textbf{Effect of hyperparameters.}
Finally, we analyze how $\lambda_1$ and $\lambda_2$ influence interaction modulation.
As shown in Fig.~\ref{fig:control_interactions}(b),
increasing $\lambda_1$ progressively amplifies interactions within the target order range, 
while increasing $\lambda_2$ produces correspondingly stronger suppression.
This confirms that the magnitudes of $\lambda_1$ and $\lambda_2$ directly govern the intensity of modulation.

\subsection{How Interaction Orders Shapes DNNs' Macroscopic Representation Capacity}
\label{subsec:Comparative Study}
In the previous subsection, we introduced modulation losses that allow DNNs to be trained with an intended emphasis on specific interaction orders.
Using these tools, we now conduct a systematic comparative study to examine how different interaction regimes shape key macroscopic representation capacity of DNNs, including structural modeling capacity, fitting versus generalization behavior, and adversarial robustness.
The results are summarized in Table~\ref{tab:comprehensive_comparison_multi_order}.
\\

\begin{figure*}[t]\centering
\includegraphics[width = 1.0 \textwidth]{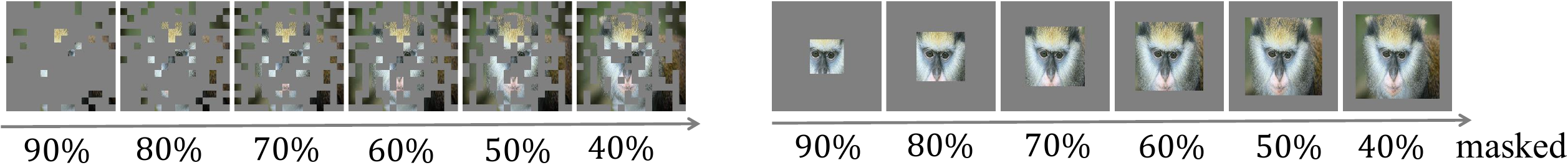}
\caption{
Illustration of the two controlled masking schemes used to probe structural sensitivity:
Random masking (left), where pixels are randomly occluded at varying ratios, disrupting global spatial structure; 
and Surrounding masking (right), where only peripheral regions are removed, preserving the central object layout.
}
\label{fig:local_global}
\end{figure*}

\begin{figure*}[htp]\centering
\includegraphics[width = 1.0 \textwidth]{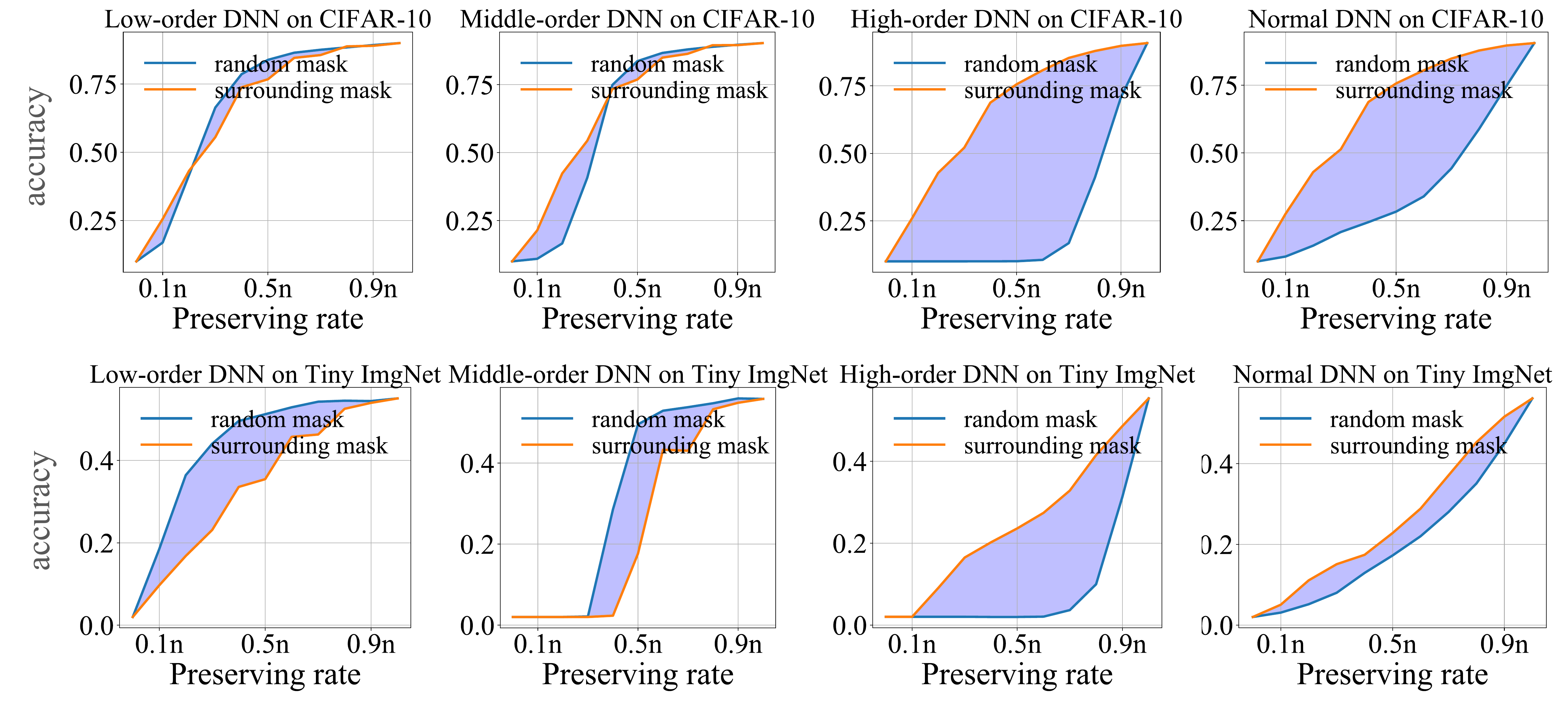}
\vspace{-12pt}
\caption{
Classification accuracy under two controlled masking perturbations for different DNN types.
Each plot compares performance on random masking versus surrounding masking across preserving ratios ($x$-axis).
The area between the two curves—the integrated accuracy difference—measures the model’s structural sensitivity.
Results are shown for four DNN types using VGG-16 trained on CIFAR-10 or TinyImageNet.
}
\label{fig:structure_lowmidhigh_vgg16}
\vspace{-4pt}
\end{figure*}

\noindent \textbf{Training four  representative types of DNNs.}
To enable a systematic comparison, we construct four representative DNN variants, each biased toward a different interaction regime.
A conventionally trained model serves as the \textit{baseline}, while three additional variants are trained using our modulation losses to predominantly encode low-order, mid-order, or high-order interactions.
We refer to these four types as \textit{normal DNNs}, \textit{low-order DNNs}, \textit{mid-order DNNs}, and \textit{high-order DNNs}.

Specifically, the normal DNNs are trained without any modulation term, i.e., $\lambda_1 = \lambda_2 = 0$.
The low-order DNNs suppress high-order interactions (within $[0.7n, n]$) by applying the suppressing loss $L^-(0.7, 1.0)$ with $\lambda_1=0$, $\lambda_2=1$.
The mid-order DNNs enhance interactions in the middle range ($[0.3n, 0.7n]$) using the encouraging loss $L^+(0.3, 0.7)$ with $\lambda_1=1$, $\lambda_2=0$.
The high-order DNNs suppress low-order interactions (within $[0, 0.5n]$) by applying $L^-(0, 0.5)$ with $\lambda_1=0$, $\lambda_2=1$.
This setup allows a controlled and direct comparison of DNNs dominated by different interaction orders.

Figure~\ref{fig:investigate_appendix} verifies that the four DNN variants indeed acquire the intended interaction structures.
Low-order DNNs markedly suppress interactions in $[0.7n, n]$, yielding distributions dominated by low-order terms.
Mid-order DNNs strongly amplify interactions in $[0.3n, 0.7n]$, producing a peak centered within this interval.
High-order DNNs suppress low-order interactions in $[0, 0.5n]$, leading to distributions concentrated at higher orders.
\vspace{4pt}

\subsubsection{Comparing structural modeling capacity}
High-order interactions naturally encode global, multi-part dependencies, whereas low-order interactions primarily capture local, part-level cues.
Thus, we hypothesize that high-order DNNs should possess the strongest structural modeling ability, while low-order DNNs should rely minimally on global structure.
To test this hypothesis, we train VGG-16 and AlexNet on CIFAR-10 and Tiny-ImageNet under each interaction configuration.

To assess structural sensitivity, we design two controlled masking settings  (Fig.~\ref{fig:local_global}):

\noindent (i) \textit{Random Masking:} randomly occludes image regions, disrupting global spatial organization and largely preserving only local, unstructured evidence.

\noindent (ii) \textit{Surrounding Masking:} masks only peripheral regions, preserving the central region that maintains the object’s overall contour and global layout.

If a model performs substantially better under surrounding masking than random masking, it indicates stronger reliance on global structural cues.
Accordingly, the performance gap between the two settings directly quantifies structural modeling capacity.
In Fig.~\ref{fig:structure_lowmidhigh_vgg16}, this gap is visualized as the area between the blue (random masking) and orange (surrounding masking) accuracy curves: larger areas correspond to greater dependence on global structure and thus richer structural representations. More experimental results are provided in supplementary materials.

\begin{figure*}[t]\centering
\includegraphics[width = 0.95 \textwidth]{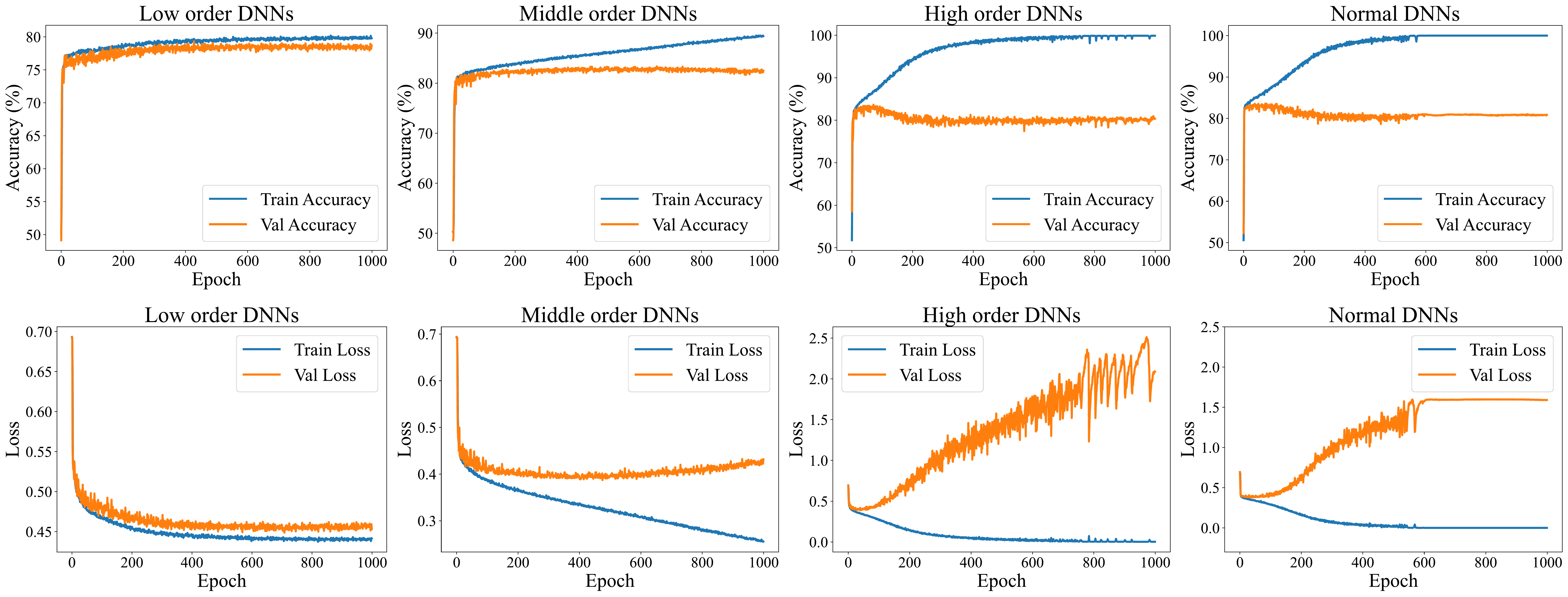}
\vspace{-4pt}
\caption{Training–testing dynamics of low-order, mid-order, high-order, and normal DNNs (three-layer MLP-8) on the Commercial dataset.
Top: training vs. validation accuracy; bottom: training vs. validation loss.
The four model variants exhibit distinct fitting and generalization behaviors consistent with their interaction-order biases. Additional results appear in the supplementary material.
}
\label{fig:fitingvsgeneralization}
\vspace{-4pt}
\end{figure*}

The results reveal three clear trends:

\noindent (i) \textbf{Low-order DNNs show minimal structural dependence.} 
Their accuracy under random masking is comparable to—or even exceeds—that under surrounding masking, indicating little reliance on global structure.

\noindent (ii) \textbf{High-order DNNs encode the richest structural information.} 
They achieve substantially higher accuracy under surrounding masking, confirming that high-order interactions capture global spatial cues.

\noindent (iii) \textbf{Mid-order and normal DNNs fall in between.}
Their accuracy gaps are moderate, reflecting partial but not dominant use of global structural information.
\vspace{4pt}

\subsubsection{Comparing Fitting and Generalization Abilities}
We next examine how different interaction orders affect a model’s fitting and generalization capacity.
Across all datasets and architectures, we observe a strikingly consistent pattern: 
low-order DNNs fit less but generalize better, whereas high-order DNNs fit nearly perfectly but generalize poorly.

To rigorously validate this phenomenon, we conduct controlled experiments using three-layer MLPs trained on the UCI Census and Commercial tabular datasets.
For each dataset, we construct four variants—normal, low-order, mid-order, and high-order DNNs—and compare their training–testing dynamics.
Figure~\ref{fig:fitingvsgeneralization} illustrates these characteristic behaviors for MLP-8 on the Commercial dataset; additional results appear in the supplementary material.
The results consistently reveal the following conclusion:

\noindent (i) \textbf{Fitting ability on training data (Low $<$ Mid/ Normal $<$ High):}
Low-order DNNs show the weakest fitting ability:
their training accuracy are substantially lower than other models, and training loss decreases only moderately (Fig.~\ref{fig:fitingvsgeneralization}, left column).
In contrast, high-order DNNs almost perfectly fit the training set—their training accuracy approaches 100\% and their training loss converges nearly to zero (right column).
Mid-order and normal DNNs fall between these two extremes.
\vspace{2pt}

\noindent (ii) \textbf{Generalization on test data (Low $>$ Mid/Normal $>$ High):}
Generalization is assessed via the gap between training and validation accuracy/loss.
Low-order DNNs exhibit nearly overlapping training and validation curves, indicating minimal generalization error (left column).
High-order DNNs show the opposite trend: as training loss approaches zero, validation loss diverges sharply upward and validation accuracy stagnates around a low plateau (right column), reflecting severe overfitting.
Again, mid-order and normal DNNs lie between the extremes.
\vspace{2pt}

\noindent (iii) \textbf{Trade-off in test accuracy:}
Despite the striking differences in fitting and generalization behaviors across the four models, their final test accuracies remain surprisingly close.
This occurs because interaction orders drive the model toward opposite inductive biases:
high-order interactions amplify fitting but weaken generalization, whereas low-order interactions strengthen generalization but limit fitting.
The test accuracy therefore emerges as the equilibrium point of these competing forces, with each model occupying a different position along the same fitting–generalization trade-off.

We further validate these findings on AlexNet and VGG-16 using CIFAR-10 and Tiny-ImageNet.
As shown in Table~\ref{tab:fitting_generalization_all}, the same pattern robustly reappears across architectures and datasets:
low-order DNNs consistently exhibit much higher training losses but the smallest loss gaps, indicating weak fitting yet strong generalization.
High-order DNNs show the opposite trend: their training losses are nearly zero but their loss gaps are the largest, reflecting severe overfitting.
Taken together, these results identify interaction order as a dominant factor governing the trade-off between fitting and generalization.
\vspace{4pt}

\begin{table*}[t]
\centering
\caption{Comparison of fitting capacity (training loss; lower is better) and generalization capacity (train–test loss gap; lower indicates better) for low-order, normal, and high-order DNNs.
}
\label{tab:fitting_generalization}
\renewcommand{\arraystretch}{1.1}
\begin{tabular}{c|cc|cc|cc|cc}
\toprule
\multirow{3}{*}{\textbf{Model}} &
\multicolumn{4}{c|}{\textbf{Fitting Capacity (Training Loss)}} &
\multicolumn{4}{c}{\textbf{Generalization Capacity (Loss Gap)}} \\
\cline{2-9}
& \multicolumn{2}{c|}{CIFAR-10} & \multicolumn{2}{c|}{TinyImageNet} &
\multicolumn{2}{c|}{CIFAR-10} & \multicolumn{2}{c}{TinyImageNet} \\
& AlexNet & VGG16 & AlexNet & VGG16 & AlexNet & VGG16 & AlexNet & VGG16 \\
\midrule
Low-order DNNs    & 0.093 & 0.059 & 0.188 & 0.226 & \textbf{0.384} & \textbf{0.396} & \textbf{1.66} & \textbf{1.75} \\
Normal DNNs       & \textbf{0.004} & \textbf{0.002} & \textbf{0.016} & \textbf{0.006} & 0.580 & 0.507 & 2.76 & 2.98 \\
High-order DNNs   & \textbf{0.004} & \textbf{0.002} & 0.027 & 0.011 & 0.761 & 0.614 & 2.88 & 3.35 \\
\bottomrule
\end{tabular}
  \label{tab:fitting_generalization_all}
\end{table*}

\subsubsection{Comparing Adversarial Robustness}
We further examine how interaction order affects adversarial robustness.
Experiments are conducted on the Census and Commercial tabular datasets using MLP-5 and MLP-8 architectures.
For each dataset, we obtain four variants—normal, low-order, mid-order, and high-order DNNs.

Adversarial examples are generated using the untargeted PGD attack \cite{madry2018towards} 
under the $L_{\infty}$ constraint $|\Delta x|_{\infty}\le\epsilon$.
We adopt $\epsilon=0.6$ (100 steps) for Census and $\epsilon=0.2$ (50 steps) for Commercial, with step size fixed at 0.01.

Table~\ref{tab:robustness_all} reports the resulting adversarial accuracies and reveals three consistent observations:

\noindent (i) \textbf{Low-order DNNs exhibit the strongest adversarial robustness:}
They surpass normal DNNs across all settings, showing that suppressing high-order interactions enhances resistance to adversarial perturbations.

\noindent  (ii) \textbf{High-order DNNs are the most vulnerable.}  
Their adversarial accuracies drop sharply, indicating that reliance on high-order interactions substantially increases attack susceptibility.

\noindent  (iii) \textbf{Mid-order and normal DNNs lie between these two extremes.}  
Their robustness levels lie between those of low-order and high-order DNNs, aligning with their intermediate structural modeling and generalization behavior.

These results support a key conclusion:
adversarial robustness is largely supported by low-order interactions, 
whereas adversarial vulnerability primarily stems from high-order interactions.
This aligns with Ren et al. \cite{ren2021game}, who observed that adversarial attacks predominantly disrupt high-order interactions.
\vspace{4pt}

\begin{table}[t]
\renewcommand{\arraystretch}{1.1}
\setlength{\tabcolsep}{3.5pt}
\caption{Comparison of adversarial robustness (adversarial accuracy, higher is better) among normal, low-order, mid-order, and high-order DNNs trained on the Census and Commercial datasets.}
\begin{tabular}{lcccc}
\toprule
\textbf{Dataset} & \multicolumn{2}{c}{\textbf{Census}} & \multicolumn{2}{c}{\textbf{Commercial}} \\
\cmidrule(lr){2-3} \cmidrule(lr){4-5}
\textbf{Model} & \textbf{MLP-5} & \textbf{MLP-8} & \textbf{MLP-5} & \textbf{MLP-8} \\
\midrule
Low-order DNNs    & \textbf{42.40} & \textbf{44.65} & \textbf{29.76} & \textbf{28.86} \\
Normal DNNs       & 38.22 & 39.33 & 27.01 & 25.92 \\
Mid-order DNNs & 15.93 & 18.10 & 23.70 & 22.55 \\
High-order DNNs   & 7.31 & 2.02 & 22.00 & 20.58 \\
\bottomrule
\end{tabular}
\label{tab:robustness_all}
\vspace{-6pt}
\end{table}

\subsubsection{Summary}
Taken together, comparative analyses reveal a clear micro–macro linkage: 
the micro-level interaction patterns that a DNN encode directly shape its macroscopic capabilities, from structural expressiveness to fitting, generalization, and adversarial robustness.

As summarized in Table~\ref{tab:comprehensive_comparison_multi_order}, each interaction regime yields a distinct and complementary capability profile:
\begin{itemize} 
    \item  \textbf{Low-order DNNs} rely on simple, localized interactions, which yield strong generalization and adversarial robustness, but limited structural expressiveness and fitting ability. 
    \item \textbf{High-order DNNs} depend on complex, context-dependent interactions, and exhibit strong structural expressiveness and fitting capacity, yet make the model fragile, thereby producing poor generalization and weak adversarial robustness.
    \item \textbf{Mid-order DNNs} fall between these two extremes. 
    Their reliance on moderately complex interactions leads to intermediate levels of structure modeling, fitting, generalization, and robustness.
\end{itemize}

Importantly, this interaction-order perspective also explains why standard DNNs work well in practice:
although not explicitly optimized for any specific order, they naturally combine the stability of low-order interactions with the expressiveness of high-order ones, leading to a balanced capability profile.

\section{Conclusion and future work}
This work establishes interaction complexity as a fundamental perspective for understanding and shaping the behavior of DNNs. 
We uncover a universal interaction bottleneck shared across architectures and tasks: 
DNNs naturally prioritize low-order and high-order interactions while systematically underrepresenting mid-order ones. 
Our theoretical analysis attributes this phenomenon to the high contextual variability of mid-order interactions, which yields unstable gradients and weak learning strength.

To overcome this bottleneck, we introduce two loss functions that explicitly modulate interaction orders. 
Furthermore, by training models dominated by low-order, mid-order, and high-order interactions, we reveal a clear micro–macro linkage: 
different interaction regimes give rise to distinct macroscopic capability profiles. 
Low-order DNNs generalize and defend against adversarial perturbations well but have limited structural modeling and fitting capacity; 
high-order DNNs excel at structural modeling and fitting but exhibit poor generalization and robustness; 
mid-order DNNs lie between these extremes. 
Together, these results show that interaction order is a key factor that shapes how a network balances fitting ability, generalization, robustness, and structural modeling.


\bibliographystyle{TPAMI_bottleneck}
\bibliography{TPAMI_bottleneck}

@inproceedings{
deng2022discovering,
title={Discovering and Explaining the Representation Bottleneck of DNNs},
author={Huiqi Deng and Qihan Ren and Hao Zhang and Quanshi Zhang},
booktitle={International Conference on Learning Representations},
year={2022},
url={https://openreview.net/forum?id=iRCUlgmdfHJ}
}

@article{yang2019xlnet,
  title={Xlnet: Generalized autoregressive pretraining for language understanding},
  author={Yang, Zhilin and Dai, Zihang and Yang, Yiming and Carbonell, Jaime and Salakhutdinov, Russ R and Le, Quoc V},
  journal={Advances in neural information processing systems},
  volume={32},
  year={2019}
}

@article{zhang2015character,
  title={Character-level convolutional networks for text classification},
  author={Zhang, Xiang and Zhao, Junbo and LeCun, Yann},
  journal={Advances in neural information processing systems},
  volume={28},
  year={2015}
}

@inproceedings{socher2013recursive,
  title={Recursive deep models for semantic compositionality over a sentiment treebank},
  author={Socher, Richard and Perelygin, Alex and Wu, Jean and Chuang, Jason and Manning, Christopher D and Ng, Andrew Y and Potts, Christopher},
  booktitle={Proceedings of the 2013 conference on empirical methods in natural language processing},
  pages={1631--1642},
  year={2013}
}

@article{warstadt2019neural,
  title={Neural network acceptability judgments},
  author={Warstadt, Alex and Singh, Amanpreet and Bowman, Samuel R},
  journal={Transactions of the Association for Computational Linguistics},
  volume={7},
  pages={625--641},
  year={2019},
  publisher={MIT Press One Rogers Street, Cambridge, MA 02142-1209, USA journals-info~…}
}

@article{lan2019albert,
  title={Albert: A lite bert for self-supervised learning of language representations},
  author={Lan, Zhenzhong and Chen, Mingda and Goodman, Sebastian and Gimpel, Kevin and Sharma, Piyush and Soricut, Radu},
  journal={arXiv preprint arXiv:1909.11942},
  year={2019}
}

@article{sanh2019distilbert,
  title={DistilBERT, a distilled version of BERT: smaller, faster, cheaper and lighter},
  author={Sanh, Victor and Debut, Lysandre and Chaumond, Julien and Wolf, Thomas},
  journal={arXiv preprint arXiv:1910.01108},
  year={2019}
}

@inproceedings{devlin2019bert,
  title={Bert: Pre-training of deep bidirectional transformers for language understanding},
  author={Devlin, Jacob and Chang, Ming-Wei and Lee, Kenton and Toutanova, Kristina},
  booktitle={Proceedings of the 2019 conference of the North American chapter of the association for computational linguistics: human language technologies, volume 1 (long and short papers)},
  pages={4171--4186},
  year={2019}
}

@article{yi2016scalable,
  title={A scalable active framework for region annotation in 3d shape collections},
  author={Yi, Li and Kim, Vladimir G and Ceylan, Duygu and Shen, I-Chao and Yan, Mengyan and Su, Hao and Lu, Cewu and Huang, Qixing and Sheffer, Alla and Guibas, Leonidas},
  journal={ACM Transactions on Graphics (ToG)},
  volume={35},
  number={6},
  pages={1--12},
  year={2016},
  publisher={ACM New York, NY, USA}
}

@inproceedings{wu20153d,
  title={3d shapenets: A deep representation for volumetric shapes},
  author={Wu, Zhirong and Song, Shuran and Khosla, Aditya and Yu, Fisher and Zhang, Linguang and Tang, Xiaoou and Xiao, Jianxiong},
  booktitle={Proceedings of the IEEE conference on computer vision and pattern recognition},
  pages={1912--1920},
  year={2015}
}

@article{qi2017pointnet++,
  title={Pointnet++: Deep hierarchical feature learning on point sets in a metric space},
  author={Qi, Charles Ruizhongtai and Yi, Li and Su, Hao and Guibas, Leonidas J},
  journal={Advances in neural information processing systems},
  volume={30},
  year={2017}
}

@inproceedings{qi2017pointnet,
  title={Pointnet: Deep learning on point sets for 3d classification and segmentation},
  author={Qi, Charles R and Su, Hao and Mo, Kaichun and Guibas, Leonidas J},
  booktitle={Proceedings of the IEEE conference on computer vision and pattern recognition},
  pages={652--660},
  year={2017}
}

@inproceedings{wu2019pointconv,
  title={Pointconv: Deep convolutional networks on 3d point clouds},
  author={Wu, Wenxuan and Qi, Zhongang and Fuxin, Li},
  booktitle={Proceedings of the IEEE/CVF Conference on computer vision and pattern recognition},
  pages={9621--9630},
  year={2019}
}

@article{wang2019dynamic,
  title={Dynamic graph cnn for learning on point clouds},
  author={Wang, Yue and Sun, Yongbin and Liu, Ziwei and Sarma, Sanjay E and Bronstein, Michael M and Solomon, Justin M},
  journal={ACM Transactions on Graphics (tog)},
  volume={38},
  number={5},
  pages={1--12},
  year={2019},
  publisher={Acm New York, NY, USA}
}

@article{han2021transformer,
  title={Transformer in transformer},
  author={Han, Kai and Xiao, An and Wu, Enhua and Guo, Jianyuan and Xu, Chunjing and Wang, Yunhe},
  journal={Advances in neural information processing systems},
  volume={34},
  pages={15908--15919},
  year={2021}
}

@inproceedings{liu2021swin,
  title={Swin transformer: Hierarchical vision transformer using shifted windows},
  author={Liu, Ze and Lin, Yutong and Cao, Yue and Hu, Han and Wei, Yixuan and Zhang, Zheng and Lin, Stephen and Guo, Baining},
  booktitle={Proceedings of the IEEE/CVF international conference on computer vision},
  pages={10012--10022},
  year={2021}
}

@article{wang2022pvt,
  title={Pvt v2: Improved baselines with pyramid vision transformer},
  author={Wang, Wenhai and Xie, Enze and Li, Xiang and Fan, Deng-Ping and Song, Kaitao and Liang, Ding and Lu, Tong and Luo, Ping and Shao, Ling},
  journal={Computational visual media},
  volume={8},
  number={3},
  pages={415--424},
  year={2022},
  publisher={Springer}
}

@inproceedings{touvron2021training,
  title={Training data-efficient image transformers \& distillation through attention},
  author={Touvron, Hugo and Cord, Matthieu and Douze, Matthijs and Massa, Francisco and Sablayrolles, Alexandre and J{\'e}gou, Herv{\'e}},
  booktitle={International conference on machine learning},
  pages={10347--10357},
  year={2021},
  organization={PMLR}
}

@article{dosovitskiy2020image,
  title={An image is worth 16x16 words: Transformers for image recognition at scale},
  author={Dosovitskiy, Alexey and Beyer, Lucas and Kolesnikov, Alexander and Weissenborn, Dirk and Zhai, Xiaohua and Unterthiner, Thomas and Dehghani, Mostafa and Minderer, Matthias and Heigold, Georg and Gelly, Sylvain and others},
  journal={arXiv preprint arXiv:2010.11929},
  year={2020}
}

@inproceedings{szegedy2015going,
  title={Going deeper with convolutions},
  author={Szegedy, Christian and Liu, Wei and Jia, Yangqing and Sermanet, Pierre and Reed, Scott and Anguelov, Dragomir and Erhan, Dumitru and Vanhoucke, Vincent and Rabinovich, Andrew},
  booktitle={Proceedings of the IEEE conference on computer vision and pattern recognition},
  pages={1--9},
  year={2015}
}

@article{zhang2022proving,
  title={Proving common mechanisms shared by twelve methods of boosting adversarial transferability},
  author={Zhang, Quanshi and Wang, Xin and Ren, Jie and Cheng, Xu and Lin, Shuyun and Wang, Yisen and Zhu, Xiangming},
  journal={arXiv preprint arXiv:2207.11694},
  year={2022}
}

@inproceedings{ren2023defining,
  title={Defining and quantifying the emergence of sparse concepts in dnns},
  author={Ren, Jie and Li, Mingjie and Chen, Qirui and Deng, Huiqi and Zhang, Quanshi},
  booktitle={Proceedings of the IEEE/CVF conference on computer vision and pattern recognition},
  pages={20280--20289},
  year={2023}
}

@inproceedings{zhou2024explaining,
  title={Explaining generalization power of a dnn using interactive concepts},
  author={Zhou, Huilin and Zhang, Hao and Deng, Huiqi and Liu, Dongrui and Shen, Wen and Chan, Shih-Han and Zhang, Quanshi},
  booktitle={Proceedings of the AAAI Conference on Artificial Intelligence},
  volume={38},
  number={15},
  pages={17105--17113},
  year={2024}
}

@article{chatterjee2022generalization,
  title={On the generalization mystery in deep learning},
  author={Chatterjee, Satrajit and Zielinski, Piotr},
  journal={arXiv preprint arXiv:2203.10036},
  year={2022}
}

@article{ilyas2019adversarial,
  title={Adversarial examples are not bugs, they are features},
  author={Ilyas, Andrew and Santurkar, Shibani and Tsipras, Dimitris and Engstrom, Logan and Tran, Brandon and Madry, Aleksander},
  journal={Advances in neural information processing systems},
  volume={32},
  year={2019}
}

@article{neyshabur2017exploring,
  title={Exploring generalization in deep learning},
  author={Neyshabur, Behnam and Bhojanapalli, Srinadh and McAllester, David and Srebro, Nati},
  journal={Advances in neural information processing systems},
  volume={30},
  year={2017}
}

@article{hochreiter1997flat,
  title={Flat minima},
  author={Hochreiter, Sepp and Schmidhuber, J{\"u}rgen},
  journal={Neural computation},
  volume={9},
  number={1},
  pages={1--42},
  year={1997},
  publisher={MIT Press One Rogers Street, Cambridge, MA 02142-1209, USA journals-info~…}
}

@inproceedings{keskar2016large,
  title={On large-batch training for deep learning: Generalization gap and sharp minima},
  author={Keskar, Nitish Shirish and Mudigere, Dheevatsa and Nocedal, Jorge and Smelyanskiy, Mikhail and Tang, Ping Tak Peter},
  booktitle={International Conference on Learning Representations},
  year={2017}
}

@article{li2018visualizing,
  title={Visualizing the loss landscape of neural nets},
  author={Li, Hao and Xu, Zheng and Taylor, Gavin and Studer, Christoph and Goldstein, Tom},
  journal={Advances in neural information processing systems},
  volume={31},
  year={2018}
}

@article{zhang2016architectural,
  title={Architectural complexity measures of recurrent neural networks},
  author={Zhang, Saizheng and Wu, Yuhuai and Che, Tong and Lin, Zhouhan and Memisevic, Roland and Salakhutdinov, Russ R and Bengio, Yoshua},
  journal={Advances in neural information processing systems},
  volume={29},
  year={2016}
}

@inproceedings{raghu2017expressive,
  title={On the expressive power of deep neural networks},
  author={Raghu, Maithra and Poole, Ben and Kleinberg, Jon and Ganguli, Surya and Sohl-Dickstein, Jascha},
  booktitle={international conference on machine learning},
  pages={2847--2854},
  year={2017},
  organization={PMLR}
}

@inproceedings{arora2016understanding,
  title={Understanding deep neural networks with rectified linear units},
  author={Arora, Raman and Basu, Amitabh and Mianjy, Poorya and Mukherjee, Anirbit},
  booktitle={International Conference on Learning Representations},
  year={2018}
}

@inproceedings{ren2022towards,
  title={Towards theoretical analysis of transformation complexity of ReLU DNNs},
  author={Ren, Jie and Li, Mingjie and Zhou, Meng and Chan, Shih-Han and Zhang, Quanshi},
  booktitle={International Conference on Machine Learning},
  pages={18537--18558},
  year={2022},
  organization={PMLR}
}

@inproceedings{yao2023towards,
  title={Towards Understanding the Generalization of Deepfake Detectors from a Game-Theoretical View},
  author={Yao, Kelu and Wang, Jin and Diao, Boyu and Li, Chao},
  booktitle={Proceedings of the IEEE/CVF International Conference on Computer Vision},
  pages={2031--2041},
  year={2023}
}

@article{shen2021interpreting,
  title={Interpreting representation quality of dnns for 3d point cloud processing},
  author={Shen, Wen and Ren, Qihan and Liu, Dongrui and Zhang, Quanshi},
  journal={Advances in Neural Information Processing Systems},
  volume={34},
  pages={8857--8870},
  year={2021}
}

@article{zhang2020game,
  title={Game-theoretic interactions of different orders},
  author={Zhang, Hao and Cheng, Xu and Chen, Yiting and Zhang, Quanshi},
  journal={arXiv preprint arXiv:2010.14978},
  year={2020}
}

@phdthesis{harsanyi1958bargaining,
  title={A bargaining model for the cooperative n-person game},
  author={Harsanyi, John C},
  year={1958},
  school={Department of Economics, Stanford University Stanford, CA, USA}
}

@inproceedings{wang2020unified,
  title={A unified approach to interpreting and boosting adversarial transferability},
  author={Wang, Xin and Ren, Jie and Lin, Shuyun and Zhu, Xiangming and Wang, Yisen and Zhang, Quanshi},
  booktitle={International Conference on Learning Representations},
  year={2021}
}

@article{pascanu2013number,
  title={On the number of response regions of deep feed forward networks with piece-wise linear activations},
  author={Pascanu, Razvan and Montufar, Guido and Bengio, Yoshua},
  journal={arXiv preprint arXiv:1312.6098},
  year={2013}
}

@article{grabisch1999axiomatic,
  title={An axiomatic approach to the concept of interaction among players in cooperative games},
  author={Grabisch, Michel and Roubens, Marc},
  journal={International Journal of game theory},
  volume={28},
  number={4},
  pages={547--565},
  year={1999},
  publisher={Springer}
}

@inproceedings{montufar2014number,
  title={On the number of linear regions of deep neural networks},
  author={Mont{\'u}far, Guido and Pascanu, Razvan and Cho, Kyunghyun and Bengio, Yoshua},
  booktitle={Advances in Neural Information Processing Systems},
  year={2014}
}

@inproceedings{sundararajan2020shapley,
  title={The shapley taylor interaction index},
  author={Sundararajan, Mukund and Dhamdhere, Kedar and Agarwal, Ashish},
  booktitle={International Conference on Machine Learning},
  pages={9259--9268},
  year={2020},
  organization={PMLR}
}

@article{shapley1951notes,
  title={Notes on the n-Person Game—II: The Value of an n-Person Game, The RAND Corporation, The RAND Corporation},
  author={Shapley, LS},
  journal={Research Memorandum},
  volume={670},
  year={1951}
}

@article{ILSVRC15,
Author = {Olga Russakovsky and Jia Deng and Hao Su and Jonathan Krause and Sanjeev Satheesh and Sean Ma and Zhiheng Huang and Andrej Karpathy and Aditya Khosla and Michael Bernstein and Alexander C. Berg and Li Fei-Fei},
Title = {{ImageNet Large Scale Visual Recognition Challenge}},
Year = {2015},
journal   = {International Journal of Computer Vision (IJCV)},
doi = {10.1007/s11263-015-0816-y},
volume={115},
number={3},
pages={211-252}
}

@article{le2015tiny,
  title={Tiny imagenet visual recognition challenge},
  author={Le, Ya and Yang, Xuan},
  journal={CS 231N},
  volume={7},
  number={7},
  pages={3},
  year={2015}
}

@article{krizhevsky2009learning,
  title={Learning multiple layers of features from tiny images},
  author={Krizhevsky, Alex and Hinton, Geoffrey and others},
  year={2009},
  publisher={Citeseer}
}

@inproceedings{krizhevsky2012imagenet,
  title={Imagenet classification with deep convolutional neural networks},
  author={Krizhevsky, Alex and Sutskever, Ilya and Hinton, Geoffrey E},
  booktitle={Advances in Neural Information Processing Systems},
  volume={25},
  pages={1097--1105},
  year={2012}
}

@inproceedings{simonyan2014very,
  title={Very deep convolutional networks for large-scale image recognition},
  author={Simonyan, Karen and Zisserman, Andrew},
  booktitle={International Conference on Learning Representations},
  year={2014}
}

@inproceedings{he2016deep,
  title={Deep residual learning for image recognition},
  author={He, Kaiming and Zhang, Xiangyu and Ren, Shaoqing and Sun, Jian},
  booktitle={Proceedings of the IEEE conference on computer vision and pattern recognition},
  pages={770--778},
  year={2016}
}

@article{lengerich2020dropout,
  title={On dropout, overfitting, and interaction effects in deep neural networks},
  author={Lengerich, Benjamin and Xing, Eric P and Caruana, Rich},
  journal={arXiv preprint arXiv:2007.00823},
  year={2020}
}

@inproceedings{zhang2020interpreting,
  title={Interpreting and boosting dropout from a game-theoretic view},
  author={Zhang, Hao and Li, Sen and Ma, Yinchao and Li, Mingjie and Xie, Yichen and Zhang, Quanshi},
  booktitle={International Conference on Learning Representations},
  year={2020}
}

@article{fort2019stiffness,
  title={Stiffness: A new perspective on generalization in neural networks},
  author={Fort, Stanislav and Nowak, Pawe{\l} Krzysztof and Jastrzebski, Stanislaw and Narayanan, Srini},
  journal={arXiv preprint arXiv:1901.09491},
  year={2019}
}

@article{xu2018understanding,
  title={Understanding training and generalization in deep learning by fourier analysis},
  author={Xu, Zhiqin John},
  journal={arXiv preprint arXiv:1808.04295},
  year={2018}
}

@article{weng2018evaluating,
  title={Evaluating the robustness of neural networks: An extreme value theory approach},
  author={Weng, Tsui-Wei and Zhang, Huan and Chen, Pin-Yu and Yi, Jinfeng and Su, Dong and Gao, Yupeng and Hsieh, Cho-Jui and Daniel, Luca},
  journal={arXiv preprint arXiv:1801.10578},
  year={2018}
}

@inproceedings{ren2021game,
  title={A Unified Game-Theoretic Interpretation of Adversarial Robustness},
  author={Ren, Jie and Zhang, Die and Wang, Yisen and Chen, Lu and Zhou, Zhanpeng and Cheng, Xu and Wang, Xin and Chen, Yiting and Shi, Jie and Zhang, Quanshi},
  booktitle={Advances in Neural Information Processing Systems},
  year={2021}
}

@inproceedings{wang2021interpreting,
  title={Interpreting attributions and interactions of adversarial attacks},
  author={Wang, Xin and Lin, Shuyun and Zhang, Hao and Zhu, Yufei and Zhang, Quanshi},
  booktitle={Proceedings of the IEEE/CVF International Conference on Computer Vision},
  year={2021}
}

@inproceedings{lundberg2017unified,
  title={A unified approach to interpreting model predictions},
  author={Lundberg, Scott M and Lee, Su-In},
  booktitle={Proceedings of the 31st international conference on neural information processing systems},
  pages={4768--4777},
  year={2017}
}

@inproceedings{madry2018towards,
  title={Towards deep learning models resistant to adversarial attacks},
  author={Madry, Aleksander and Makelov, Aleksandar and Schmidt, Ludwig and Tsipras, Dimitris and Vladu, Adrian},
  booktitle={International Conference on Learning Representations},
  year={2018}
}

@article{tsang2018neural,
  title={Neural interaction transparency (nit): Disentangling learned interactions for improved interpretability},
  author={Tsang, Michael and Liu, Hanpeng and Purushotham, Sanjay and Murali, Pavankumar and Liu, Yan},
  journal={Advances in Neural Information Processing Systems},
  volume={31},
  pages={5804--5813},
  year={2018}
}

@inproceedings{peebles2020hessian,
  title={The hessian penalty: A weak prior for unsupervised disentanglement},
  author={Peebles, William and Peebles, John and Zhu, Jun-Yan and Efros, Alexei and Torralba, Antonio},
  booktitle={Computer Vision--ECCV 2020: 16th European Conference, Glasgow, UK, August 23--28, 2020, Proceedings, Part VI 16},
  pages={581--597},
  year={2020},
  organization={Springer}
}

@article{novak2018sensitivity,
  title={Sensitivity and generalization in neural networks: an empirical study},
  author={Novak, Roman and Bahri, Yasaman and Abolafia, Daniel A and Pennington, Jeffrey and Sohl-Dickstein, Jascha},
  journal={arXiv preprint arXiv:1802.08760},
  year={2018}
}

\ifCLASSOPTIONcaptionsoff
  \newpage
\fi

\vspace{-30pt}
\begin{IEEEbiography}[{\includegraphics[width=1in,height=1.5in,clip,keepaspectratio]{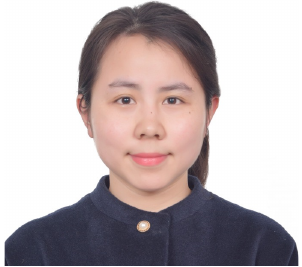}}]{Dr. Huiqi Deng} is an assistant professor at Xi'an Jiao Tong University, China. 
She received her Ph.D. degree in Applied Mathematics from Sun Yat-sen University, China, in 2021.
From 2022 to 2024, she served as a postdoctoral researcher at Shanghai Jiao Tong University, China. 
Her research focuses on various aspects of trustworthy AI, including explainability, generalization, and robustness.
To date, she has published over 20 papers in top journals and conferences, such as IEEE TPAMI, NeurIPS, ICML, ICLR, AAAI, and KDD.
\end{IEEEbiography}

\vspace{-30pt}
\begin{IEEEbiography}[{\includegraphics[width=1in,height=1.5in,clip,keepaspectratio]{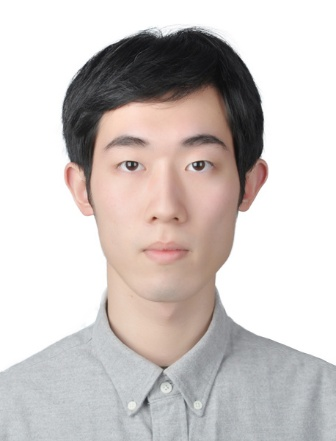}}]{Qihan Ren}  is a Ph.D. student at Shanghai Jiao Tong University. He received a bachelor’s degree in Electrical and Computer Engineering from Shanghai Jiao Tong University, and a dual bachelor’s degree in Computer Science from University of Michigan, U.S., in 2022. His research interests include explainable AI (XAI), machine learning, computer vision, and natural language processing. He has published 8 papers in top-tier venues, such as IEEE TPAMI , ICML, NeurIPS, ICLR.
\end{IEEEbiography}

\vspace{-30pt}
\begin{IEEEbiography}[{\includegraphics[width=1in,height=1.5in,clip,keepaspectratio]{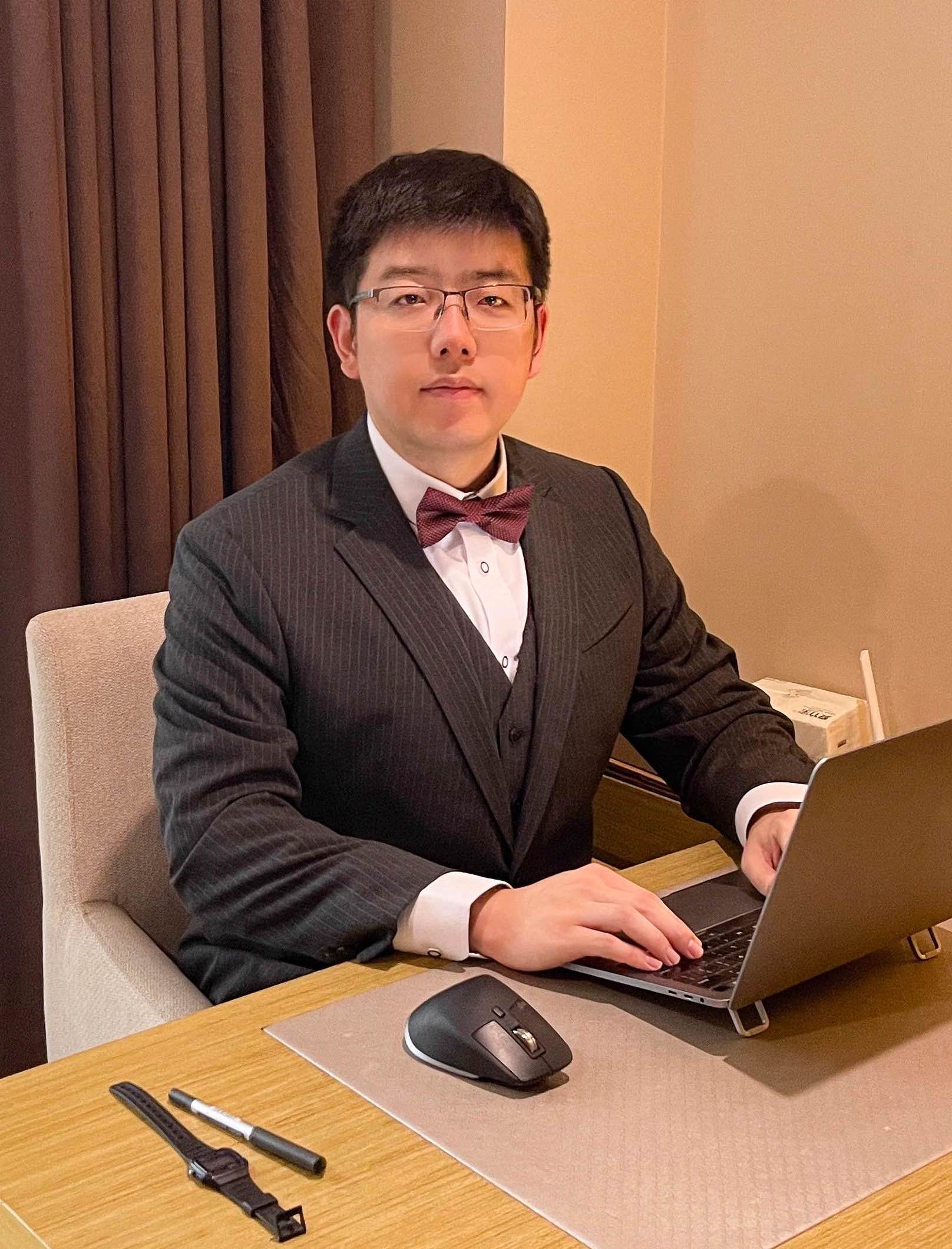}}]
{Dr. Hongbin Pei}  is an Assistant Professor at Xi’an Jiaotong University, China. He received his B.S., M.S., and Ph.D. degrees from Jilin University in 2012, 2015, and 2021, respectively. He was a visiting scholar at the University of Illinois at Urbana-Champaign (UIUC) and Hong Kong Baptist University (HKBU). His research focuses on graph learning, geometric deep learning, and spatio-temporal data mining, with applications for social good. He serves as a senior program committee member and reviewer for conferences and journals, including CVPR, ICML, ICLR, IEEE TPAMI, and TNNLS.
\end{IEEEbiography}

\vspace{-30pt}
\begin{IEEEbiography}[{\includegraphics[width=1in,height=1.5in,clip,keepaspectratio]{./figures/zhang.pdf}}]{Dr. Quanshi Zhang}
is an associate professor at Shanghai Jiao Tong University, China. He received the Ph.D. degree from the University of Tokyo in 2014. From 2014 to 2018, he was a post-doctoral researcher at the University of California, Los Angeles. His research interests are mainly machine learning and computer vision. In particular, he has made influential research in explainable AI (XAI). He won the ACM China Rising Star Award at ACM TURC 2021. He is the speaker of the tutorials on XAI at IJCAI 2020 and IJCAI 2021. He was the co-chairs of the workshops towards XAI in ICML 2021, AAAI 2019, and CVPR 2019. 
\end{IEEEbiography}

\end{document}